%% file: bare_jrnl.tex
\documentclass[journal]{IEEEtran}
\ifCLASSINFOpdf
\else
\fi

\usepackage{epsfig}
\usepackage{times}
\usepackage{bm}
\usepackage{mathrsfs}
\usepackage{array}
\usepackage{booktabs}
\usepackage{epsfig}
\usepackage{graphicx}
\usepackage{subfigure}
\usepackage{amsmath}
\usepackage{multirow}
\usepackage{amssymb}
\usepackage{algorithm}
\usepackage{algorithmic}
\usepackage{cite}
\usepackage{caption}

\begin{document}
\title{Efficient Discrete Supervised Hashing for Large-scale Cross-modal Retrieval}
%
\author{Tao Yao,
        Xiangwei Kong, ~\IEEEmembership{Member,~IEEE,} 
        Lianshan Yan,~\IEEEmembership{Senior Member,~IEEE,}
        Wenjing, Tang,
        and~Qi~Tian,~\IEEEmembership{Fellow,~IEEE}
\thanks{T. Yao is with the Department of Information and Electrical Engineering, Ludong University, Yantai, China, 264000 and Yantai Research Institute of New Generation Information Technology at Southwest Jiaotong University, Yantai, China, 264000 (email: yaotao@ldu.edu.cn). X. Kong is with the Department of Data Science and Engineering Management, Zhejiang University, China, 310058 (e-mail: kongxw@dlut.edu.cn). L. Yan is with Yantai Research Institute of New Generation Information Technology, Southwest Jiaotong University, Yantai, China, 264000 (e-mail: lsyan@home.swjtu.edu.cn). W. Tang is with the Department of Information and Electrical Engineering, Ludong University, Yantai, China, 264000 (tangwenjing@ldu.edu.cn). Q. Tian is with the Department of Computer Science, University of Texas at San Antonio, San Antonio, TX 78249 USA (e-mail: qitian@cs.utsa.edu).}
}

\maketitle
\begin{abstract}
Supervised cross-modal hashing has gained increasing research interest on large-scale retrieval task owning to its satisfactory performance and efficiency. However, it still has some challenging issues to be further studied: 1) most of them fail to well preserve the semantic correlations in hash codes because of the large heterogenous gap; 2) most of them relax the discrete constraint on hash codes, leading to large quantization error and consequent low performance; 3) most of them suffer from relatively high memory cost and computational complexity during training procedure, which makes them unscalable. In this paper, to address above issues, we propose a supervised cross-modal hashing method based on matrix factorization dubbed Efficient Discrete Supervised Hashing (EDSH). Specifically, collective matrix factorization on heterogenous features and semantic embedding with class labels are seamlessly integrated to learn hash codes. Therefore, the feature based similarities and semantic correlations can be both preserved in hash codes, which makes the learned hash codes more discriminative. Then an efficient discrete optimal algorithm is proposed to handle the scalable issue. Instead of learning hash codes bit-by-bit, hash codes matrix can be obtained directly which is more efficient. Extensive experimental results on three public real-world datasets demonstrate that EDSH produces a superior performance in both accuracy and scalability over some existing cross-modal hashing methods.
\end{abstract}
\section{Introduction}
With the volume of data increasing explosively on Internet, how to efficiently retrieve similar data points in large-scale datasets has become a challenging issue. Due to the high complexity of computing the similarities between the query and candidate data points, traditional nearest neighbor search methods are not suitable for scalable retrieval task. Therefore, how to significantly accelerate retrieval speed with low memory cost becomes an imperative requirement. In recent years, hashing method, which maps the data points from a high dimensional feature space to a low dimensional Hamming space by preserving the data structure in the original space, has attracted a lot of attention \cite{Shen2015Learning,Nonliner2017chen,Shen2015Supervised,regular2018chen,Lin2015Supervised,Yao2019Online,Chen2018Graph,Yao2016Semantic,Shen2016Semi,Liu2016Sequential,Tang2016Supervised}. For hashing methods, the similarities between the query and candidate data points can be calculated efficiently by XOR operation. Due to its efficiency in memory and computational cost, hashing method has become an important tool to handle large-scale applications, \emph{e.g.} person re-identification, classification, reconstruction, retrieval and so on \cite{Zhu2017Part,Hu2019Collective,Wu2018K,hashing2017chen}. \par

Most early hashing works only focus on one modality, \emph{e.g.} using an image to retrieve similar images \cite{Shen2015Learning,Nonliner2017chen,Shen2015Supervised,regular2018chen,Lin2015Supervised,Chen2018Graph}. However, data points typically are represented by multi-modalities in real applications, which makes today's retrieval task even more challenging. Furthermore, users often expect search engine to return similar data points with different modalities when users submit a query to search engines. However, single-modal hashing can not be extended to cross-modal hashing directly owing to the heterogeneous gap among different modalities. Specifically, different modalities lie in different feature spaces, which makes the similarities between heterogeneous data points unable to be measured directly. Last decade has witnessed continued efforts on improving the performance and efficiency of the cross-modal retrieval task, which is diffusely known as one of fundamental issue in computer vision \cite{Rasiwasia-et-al:scheme,cross2017ding,Zhang2015Full,Huang2017Cross,Peng2016Semi,Jia2011Learning,Peng2017An}.\par

Accordingly, many cross-modal hashing approaches have been designed and shown promising performance \cite{Ji-et-al:scheme,Mandal2018Generalized,Pan-et-al:scheme,Shen2015Supervised,Wang2012Semi,Lin2015Supervised,Shen2013Inductive,Zhang-et-al:scheme,Tang2016Supervised,Liu2016Structure}. In term of whether class labels are used in training procedure, cross-modal hashing methods mainly consist of two categories, \emph{i.e.}, unsupervised hashing methods and supervised ones. The former aims at learning hash functions by preserving feature based similarities in Hamming space. However, the learned sharing Hamming space is independent with class labels, which makes it suffer from less discriminative hash codes and consequent low retrieval performance. In contrast, the latter leverages class labels of training data points to learn the sharing Hamming space to improve retrieval performance. Most existing supervised methods firstly exploit class labels to construct a pairwise similarity matrix, and then hash functions are learned by preserving the pairwise similarities in Hamming space \cite{Liu2018SDMCH,Yao2016Semantic,Tang2016Supervised}. However, these approaches have two drawbacks: 1) Converting class labels to the pairwise similarity matrix results in category information loss which inevitably degrades the quality of hash codes; 2) The size of similarity matrix is too large on large-scale applications, which generally results in large memory cost and high computational complexity.\par

Another problem for supervised hashing works is how to effectively preserve the similarities in data itself in the learned Hamming space. Most of them only embed semantic correlations with class labels into hash codes learning. Nevertheless, many heterogeneous data points with same class labels are hard to be close in the sharing Hamming space since the semantic correlations in the heterogeneous data points are complex.\par

Furthermore, the optimal problems for hashing methods are hard to solve because of the discrete constraint on hash codes. Therefore, most existing methods generally discard the discrete constraint to obtain continuous solutions and then hash codes are generated by thresholding operation. However, to relax the discrete constraint generally results in large quantization error and consequent suboptimal retrieval performance. Accordingly, many discrete cross-modal hashing methods have been proposed to address this issue \cite{Liu2016Sequential,Shen2015Supervised,Shen2016Semi}. However, these methods generally learn hash codes bit-by-bit, which is time-consuming on large-scale applications. \par

To address the above challenges, we present a supervised hashing method, termed Efficient Discrete Supervised Hashing, EDSH for short. Specifically, a sharing space is learned by collective matrix factorization to reduce the heterogeneous gap firstly. And then an orthogonal matrix is learned to bridge the semantic correlations between the sharing space and Hamming space which is derived from class labels. This two-step process can make the learned hash codes more discriminative. Finally, a discrete optimal method is proposed to learn discrete hash codes directly. More significantly, hash codes can be learned in an efficient manner instead of learning them bit-by-bit. The flowchart of our EDSH is described in Fig.1. \par

In particular, the main contributions of this work are summarized as follows:\par
1) Unlike most existing works which only preserve feature based similarities or semantic correlations in hash codes, a novel supervised cross-modal hashing method is proposed to preserve both of them in hash codes, which makes hash codes more discriminative. \par
2) To enable large-scale applications, we develop an efficient discrete optimization method, which can learn hash codes rapidly. Unlike most existing discrete hashing methods which learn hash codes bit-by-bit \cite{Liu2016Sequential,Zhang-Li:kl-one,Shen2015Supervised,Shen2016Semi,Xu2017TIP}, the hash codes have a closed-form solution which leads to a rapid convergence. \par
3) We conduct experiments over three real-world multi-modal datasets to show the effectiveness and efficiency of our EDSH. The experimental results demonstrate the superiority of EDSH in both accuracy and scalability.\par
The organization of the rest part is given as follows. Section 2 reviews some related works. Section 3 introduces the proposed EDSH model. Section 4 presents extensive experimental results and corresponding analysis on three public datasets. Finally, Section 5 gives conclusion of this work.\par

\begin{figure*}[t]
\begin{center}
\includegraphics[width=18cm,keepaspectratio]{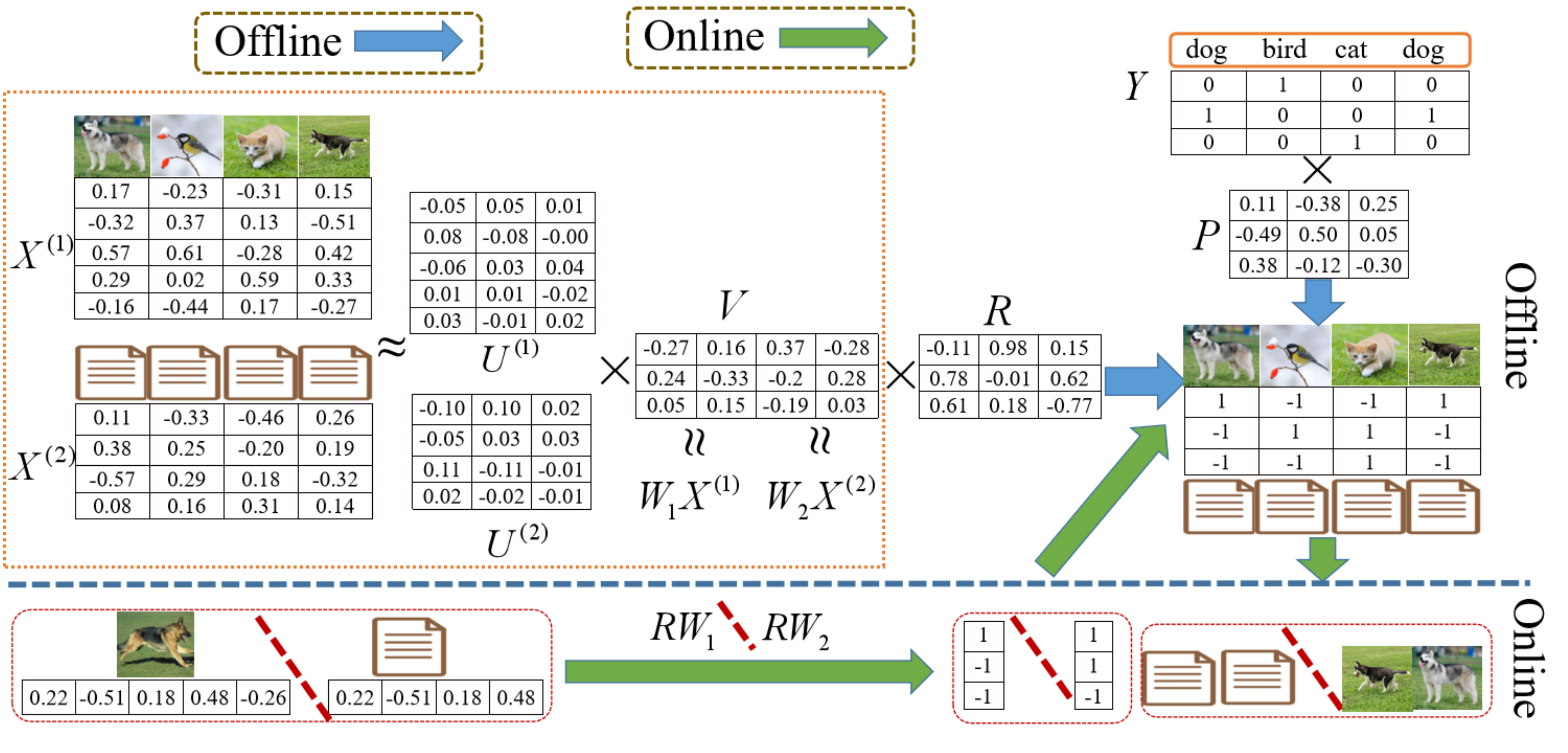}
\caption{The algorithmic flowchart of EDSH. The overall phase can be divided into two parts: online and offline. For the online part, collective matrix factorization on heterogenous features and semantic embedding with class labels are incorporated to learn hash codes where the feature based similarities and semantic correlations can be both preserved. Then hashing functions are learned to map data points to the sharing space. For the offline phase, the hash code of the query can be generated directly, then the distances between the query and candidate data points can be efficiently calculated by XOR operation. At last, those data points with relatively smaller Hamming distance are returned.}
\end{center}
\end{figure*}
\section{Related Work}
Cross-modal hashing is a classical problem in multi-view learning. Due to its effectiveness and efficiency, cross-modal hashing has drawn considerable attention. More recently, many cross-modal hashing works have been proposed. Cross-modal hashing methods can be roughly categorized into two streams: unsupervised and supervised methods. We briefly review the two kinds of works in this section.\par

\textbf{Unsupervised cross-modal hashing methods} pay attention on learning a sharing Hamming space by preserving feature based similarity without supervised information. Inter-Media Hashing (IMH) proposes to preserve inter-modal and intra-modal similarities in heterogeneous data points to learn a sharing hamming space, and then linear hash functions are learned by a linear regression model \cite{songjk2013}. Latent Semantic Sparse Hashing (LSSH) proposes to learn semantic subspaces for text and image modalities by matrix factorization and sparse coding respectively, and then a mapping matrix are learned by preserving the inter-modal similarity \cite{Zhou-et-al:scheme}. Collective Matrix Factorization Hashing (CMFH) proposes to learn a sharing subspace by preserving inter-modal similarity with collective matrix factorization firstly, and then linear hash functions are learned by a linear regression model \cite{Ding-et-al:scheme}. Fusion Similarity Hashing (FSH) formulates to explicitly capture the heterogeneous correlations in data points by preserving the fusion similarity in data points \cite{Liu2017Cross}. However, above mentioned approaches embed heterogeneous data points to a continuous isomorphic space, and then hash codes are gained by thresholding operation separately. These methods result in large quantization error, which degrades the discrimination of hash codes and substantial loss of retrieval accuracy. Composite Correlation Quantization (CCQ) proposes to learn correlation-maximal hash functions and composite quantizers jointly \cite{Long2016Composite}. Intra-modal and inter-modal similarity are preserved by minimizing both reconstruction and quantization errors. Quantized Correlation Hashing (QCH) takes both hash functions learning and quantization loss into consideration  to generate more discriminative hash codes \cite{wu2015quantized}. Alternating Co-Quantization (ACQ) proposes to alternately learn binary quantizers for each modality, which can minimize quantization errors while preserving the similarity in data points \cite{Irie2016Alternating}. However, these methods aim at learning hash functions by original features directly, which limits the retrieval performance because of the complexity correlations in heterogenous data points. Furthermore, above methods do not take class labels into consideration, which generally degrades the retrieval performance. \par

\textbf{supervised cross-modal hashing methods} aim at learning more discriminative hash codes by incorporating the supervised information to hash functions learning. Cross-Modality Similarity Sensitive Hashing (CMSSH) extends similarity-sensitive hashing to facilitate cross-modal retrieval \cite{Bronstein-et-al:scheme}. Spectral Hashing (SH) is applied to multi-modal data points, namely Cross-View Hashing (CVH) \cite{Kumar-Udupa:kl-one}. And then nonlinear hash functions are learned for out-of-samples by kernel trick. Semantic Correlation Maximization (SCM) constructs pairwise similarity matrix by class labels firstly, and then the hash codes are learned by approximating the pairwise similarity \cite{Zhang-Li:kl-one}. For efficiency, SCM proposes to learn hash codes bit-by-bit. Fast Discrete Cross-modal Hashing (FDCH) proposes to regress the class labels to learn hash codes with a drift \cite{Liu2018Fast}. Semantics Preserving Hashing (SePH) proposes to minimizes the Kullback-Leibler divergence (KLD) between pairwise similarity matrix and hash codes \cite{Lin2015Semantics}. However, the computational complexity the and memory cost of SePH increase rapidly as the training samples increasing, which limits its application on large-scale datasets. Supervised Matrix Factorization Hashing (SMFH) incorporates semantic labels to learn a sharing space based on matrix factorization, and then hash functions are learned by the sharing space \cite{Tang2016Supervised}. Label Consistent Matrix Factorization Hashing (LCMFH) maps heterogeneous data points into a latent space, and then align the latent space with a latent semantic space learned from class labels \cite{Devraj2017}. However, these methods do not take quantization loss into consideration, which degrade the discrimination of hash codes. Discrete Cross-modal Hashing (DCH) aims at learning discriminative hash codes by discrete cyclic coordinate descent (DCC) method with a bit-by-bit manner \cite{Xu2017TIP}. However, the bit-by-bit manner is still much time-consuming.\par
On the one hand, most existing supervised cross-modal hashing methods learn hash codes by preserving the pairwise similarities \cite{Lin2015Semantics,Liu2018Fast,Kumar-Udupa:kl-one,Liong2018Cross}. However, these methods either cost much computational source in training phase or fail to explicitly capture the semantic correlations in multi-modal data points. On the other hand, most existing supervised cross-modal hashing methods typically relax the discrete constraints on hash codes to obtain continuous solutions, then hash codes are generated by thresholding operation \cite{Lin2015Semantics,Liu2018Fast,Kumar-Udupa:kl-one}. They generally can not achieve promising retrieval performance because of the large quantization loss. Although some discrete hashing methods have been proposed to address this issue by learning hash codes bit-by-bit \cite{Liu2016Sequential,Xu2017TIP,Shen2016Semi}, this manner is still much time-consuming. To address above issues, in this paper, we propose a novel cross-modal hashing method to learn more discriminative hash codes directly via an efficient discrete optimal algorithm.\par
\section{Efficient Discrete Supervised Hashing}
In this section, we present the details of our EDSH. The proposed EDSH consists of four components: notations, formulation, efficient discrete optimization algorithm and computational complexity analysis, which are described in Section 3.1, 3.2, 3.3 and 3.4, respectively.
\subsection{Notations}
To simplify the presentation, supposing that each data point has two modalities, \emph{i.e.} image and text. Without loss of generality, it can be easily extended to more than two modalities. Assuming that we have $N$ data points described by $X=\{X^{(1)}, X^{(2)}\}$, and $X^{(1)} \in \mathcal{R}^{ d_1\times N}$, $X^{(2)} \in \mathcal{R}^{ d_2\times N}$, where $d_1$ and $d_2$ are the dimensions of the two modalities, respectively. $X^{(m)}=\{x^{(m)}_1, x^{(m)}_2, x^{(m)}_3, \cdots x^{(m)}_N\}$, $m=\{1,2\}$, where $x^{(m)}_i$ denotes the $i$-th data point of the $m$-th modality. Besides the two feature vectors, class labels are also available $Y=\{y_1, y_2, y_3 \cdots y_N\} \in \{0,1\}^{c\times N}$, where $c$ denotes the total number of categories, and $y_{ij}=1$ if $x_i$ belongs to the $j$-th semantic category and $0$ otherwise. Without loss of generality, we assume that the feature vectors are zero-centered, \emph{i.e.} $\sum_{i=1}^N x^{(1)}_{ij}=0$, $\sum_{i=1}^N x^{(2)}_{ij}=0$.\par
Given $N$ training data points, cross-modal hashing aims at learning two groups hash functions $W_1$ and $W_2$ to map the heterogeneous data points to a sharing Hamming space, and $B=\{-1,1\}^{k \times N}$ denotes the hash codes of training data points, where $k$ is the length of hash codes. For simplicity, the linear mapping is adopted as hash functions which are defined as following\par
\begin{equation} h^{(1)} (x^{(1)}_i)=sgn(W_1 x^{(1)}_i)\end{equation}
\begin{equation} h^{(2)} (x^{(2)}_i)=sgn(W_2 x^{(2)}_i)\end{equation}
where $sgn(\cdot)$ is an element-wise sign function.\par
\subsection{Formulation}
\subsubsection{Sharing Space Learning}
Due to the complex correlations in heterogeneous data points, it is probably hard to directly map data points with same class labels to be close in the learned sharing Hamming space. To better bridge the semantic correlations between Hamming space and original feature space, we propose to learn a sharing space for heterogeneous data points firstly. In this paper, collective matrix factorization algorithm, which can learn unified representations for pairwise heterogeneous data points, is employed to learn a sharing space by preserving inter-modal similarity. Thus, we formulate the sharing space learning issue by minimizing
\begin{equation} \mathop{\arg\min}_{U^{(m)},V}  \sum_{m=1}^{2} \lambda_m \left\| X^{(m)}-U^{(m)}V \right\|_F^2+ \mu (\left\| U^{(m)}\right\|_F^2+ \left\|V\right\|_F^2) \end{equation}
where $U^{(m)}\in \mathcal{R}^{d_m \times k}$($m=\{1, 2\}$) is a mapping matrix, and $V\in \mathcal{R}^{k \times N}$ is the sharing space. $\lambda_m$ and $\mu$ are weighted parameters.\par
\subsubsection{Hash Functions Learning}
Although the sharing space is learned in training phase for training data points, we need to learn modality-specific hash functions to address the out-of-sample issue. The hash functions can be learned by solving the following problem
\begin{equation} \mathop{\arg\min}_{W_m}  \sum_{m=1}^{2} \beta_m \left\| V- W_m X^{(m)}\right\|_F^2 +\mu \left\| W_m\right\|_F^2 \end{equation}
where $W_m\in \mathcal{R}^{k \times d_m}$($m=\{1, 2\}$) is the hash function for the $m$th modality, and $\beta_m$ is a weighted parameter.\par
\subsubsection{Hash Codes Learning}
Class labels contain high-level semantic information, which can improve the discrimination of hash codes. Accordingly, several supervised cross-modal hashing methods are proposed to firstly construct a pairwise similarity matrix derived by the class labels, then hash codes are learned by preserving the pairwise similarity in Hamming space \cite{Zhang-Li:kl-one,Lin2015Semantics}. However, the $N\times N$ similarity matrix inevitably results in large computational cost. Moreover, transforming class labels to similarity matrix results in category information loss, which may degrade the discrimination of hash codes. \par
In this paper, class labels are directly embedded into hash codes learning by assuming that the class labels have semantic correlations with hash codes, \emph{i.e.,}
\begin{equation} \mathop{\arg\min}_{P} \gamma \left\|Y - P B\right\|_F^2 \end{equation}
where $P \in \mathcal{R}^{c \times k}$ is a latent semantic space to bridge the semantic correlations between class labels and hash codes, and $\gamma$ is a weighted parameter. Each column of $P$ denoted by $p_i$ represents a latent semantic concept, and $B_{ij}=1$ denotes that the $j$-th data point contains the $i$-th latent semantic concept otherwise $B_{ij}=0$. Specifically, the high-level class labels can be reconstructed by several latent semantic concepts. That is, those data points with same class label should have same hash codes which is reasonable.\par
\subsubsection{Bridging The Correlations Between Hash Codes And The Sharing Space}
To bridging the semantic correlations between binary codes and the sharing space, we further define the following formulation:
\begin{equation} \mathop{\arg\min}_{R} \alpha \left\| B- RV\right\|_F^2 \end{equation}
$$s.t.  RR^T=I, B\in \{-1,1\}^{k\times N}$$
where $R\in \mathcal{R}^{k\times k}$ is an orthogonal rotation matrix, and $I$ denotes the identity matrix. With this formulation, discrete hash codes can be learned directly in the training procedure as shown in Section 3.3. Therefore, the quantization loss can be avoided, which makes the learned hash codes more discriminative.\par
\subsubsection{Objective Function}
Combining the collective matrix factorization term given in Eq.(3), the hash functions learning term given in Eq.(4), the hash codes learning term given in Eq.(5), the bridging the correlations between hash codes and the sharing space term given in Eq.(6) and the regularization term, the overall objective function of EDSH is defined as
\begin{align} \begin{split} \mathop{\arg\min}_{U^{(m)},V, B, R, P, W_m}  &\sum_{m=1}^{2} \lambda_m \left\| X^{(m)}-U^{(m)}V \right\|_F^2+ \gamma\left\| Y-PB \right\|_F^2 \\ &+\alpha \left\| B- RV\right\|_F^2 + \beta_m \left\| V- W_m X^{(m)}\right\|_F^2 \\& +\mu Reg( U^{(1)},U^{(2)},P,V, B, R, W_m) \end{split} \end{align}\par
$s.t.  RR^T=I, B\in \{-1,1\}^{k\times N}$\\
where $Reg(\cdot)$ denotes the regulation term to avoid overfitting, and $Reg(\cdot)=(\left\|U^{(1)}\right\|_F^2+\left\|U^{(2)}\right\|_F^2+ \left\|V\right\|_F^2+\left\|W_1\right\|_F^2+\left\|W_2\right\|_F^2)$.\par
Our objective function is formulated to learn hash codes via preserving both feature based and class label based similarities, such that the learned hash codes are optimal for cross-modal retrieval task.\par
Note that although LSSH, CMFH, SMFH, LCMFH and our EDSH are all matrix factorization based cross-modal hashing methods, our main formulation is different with other methods. Firstly, LSSH and CMFH are unsupervised methods, while the others are supervised ones which can enhance the semantic information in hash codes and consequent promising performance. Secondly, SMFH proposes to learn a sharing semantic space by incorporating the Laplacian matrix (whose size is $N \times N$) which makes it unscalable to large-scale datasets. Finally, SMFH and LCMFH propose to learn a sharing space firstly, and then hash codes can be generated by quantize the continuous representations directly. The quantization error typically degrades the discrimination of hash codes, while our EDSH can learn discrete hash codes directly.\par
\subsection{Efficient Discrete Optimization Algorithm}
Apparently, the optimization problem Eq.(7) is not convex with the matrices variables $U^{(1)}$, $U^{(2)}$, $P$, $V$, $B$, $R$, $W_1$, $W_2$, and thus hard to solve. Fortunately, it is convex with respect to any one of $U^{(1)}$, $U^{(2)}$, $P$, $V$, $B$, $R$, $W_1$, $W_2$ when other variables are fixed. Therefore, we propose an alternative optimal algorithm to solve the subproblems with respect to each variable. The details of our optimal algorithm are given as following. \par
$\bm{U^{(1)}}$ \textbf{-step:} Fixing the other variables and dropping the irrelevant terms to $U^{(1)}$, we obtain
\begin{align} \mathop{\arg\min }_{U^{(1)}} \lambda_1\left\| {X^{(1)}-U^{(1)}V} \right\|_F^2 + \mu (\left\| U^{(1)} \right\|_F^2) \end{align}\par
Letting the derivation of Eq.(7) with respect to $U^{(1)}$ equal zero, we have
\begin{align} U^{(1)}VV^T-X^{(1)}V^T- \frac{\mu}{\lambda_1} U^{(1)}=0 \end{align}\par
Therefore, we can obtain a closed-form solution for $U^{(1)}$ as following
\begin{align} U^{(1)}=X^{(1)}V^T(VV^T+\frac{\mu}{\lambda_1})^{-1} \end{align}\par
$\bm{U^{(2)}}$ \textbf{-step:} Similar to solve $U^{(1)}$, we have
\begin{align} U^{(2)}=X^{(2)}V^T(VV^T+\frac{\mu}{\lambda_2})^{-1} \end{align}\par
$\bm{P}$ \textbf{-step:} Similar to solve $U^{(1)}$, we have
\begin{align} P=Y B^T(BB^T)^{-1} \end{align}\par
$\bm{V}$ \textbf{-step:} Fixing the other variables and letting the derivation of Eq.(7) with respect to $V$ equal zero, we have
\begin{align} \begin{split}&\sum_{m=1}^{2}\lambda_m U^{(m)T}U^{(m)}V +\alpha R^TRV+ (\beta_1+\beta_2+\mu)V \\&- \sum_{m=1}^{2}\lambda_m U^{(m)T} X^{(m)}-\alpha R^TB-\beta_1 W_1X^{(1)} \\&- \beta_2 W_2X^{(2)}=0 \end{split}\end{align}\par
Therefore, we can obtain a closed-form solution for $V$ as following:
\begin{align} \begin{split}V=&(\sum_{m=1}^{2} \lambda_m U^{(m)T}U^{(m)}+\alpha R^TR + (\beta_1+\beta_2 \\&+\mu)I)^{-1}(\sum_{m=1}^{2} \lambda_m U^{(m)T} X^{(m)} + \alpha R^TB\\&+ \beta_1 W_1X^{(1)}+\beta_2 W_2X^{(2)}) \end{split}\end{align}\par

$\bm{R}$ \textbf{-step:} Fixing the other variables and dropping the irrelevant terms to $R$, we obtain
\begin{equation}\mathop{\arg\min }_{R}\left\| B-RV \right\|_F^2 \end{equation}
$$s.t. RR^T=I$$ \par

Obviously, this sub-problem is a classical Orthogonal Procrustes problem \cite{Scho-Peter:kl-one}, which can be solved by Singular Value Decomposition (SVD). Specifically, the SVD of $BV^T$ is firstly computed as $BV^T=S\Omega \hat{S}^T$, and then the orthogonal rotation matrix can be updated by $R={\hat S}S^T$.

$\bm{B}$ \textbf{-step:} Fixing the other variables and dropping the irrelevant terms to $B$, we obtain
\begin{equation}\mathop{\arg\min }_{B}\alpha \left\| B-RV \right\|_F^2 + \gamma\left\| Y-PB \right\|_F^2 \end{equation}
$$s.t. B \in \{-1,1\}^{k \times N}$$\par
Eq.(16) is equivalent to
\begin{equation} \begin{split}\mathop{\arg\min }_{B} \quad &\alpha tr(B^T B) - 2\alpha tr(V^T R^T B ) \\&+ \gamma tr(B^T P^T P B) -2\gamma tr(Y^T PB )\end{split} \end{equation}
Since $tr(B^T B)$ and $tr(B^T P^T P B)$ are constants, we can obtain a closed-form solution for $B$ as following
\begin{equation}B=sgn(\alpha RV+\gamma P^T Y) \end{equation}\par

$\bm{W_1}$ \textbf{-step:} Fixing the other variables and dropping the irrelevant terms to $W_1$, we obtain
\begin{equation}\mathop{\arg\min }_{W_1}\left\| V-W_1 X^{(1)} \right\|_F^2 + \left\|W_1\right\|_F^2  \end{equation}\par
Letting the derivation of Eq.(19) with respect to $W_1$ equal zero, we have
\begin{align} W_1X^{(1)}X^{(1)T}- VX^{(1)T}+\frac{\mu}{\beta_1} W_1=0 \end{align}\par
Therefore, we can obtain a closed-form solution for $W_1$ as following
\begin{align} W_1= V X^{(1)T}(X^{(1)}X^{(1)T}+\frac{\mu}{\beta_1}I)^{-1} \end{align}\par
$\bm{W_2}$ \textbf{-step:} Similar to solve $W_1$, we have
\begin{align} W_2= V X^{(2)T}(X^{(2)}X^{(2)T}+\frac{\mu}{\beta_2}I)^{-1} \end{align}\par
By repeating the above eight steps until it converges or the number of iterations reaches the maximum value, the optimal solutions can be obtained. It is worth noting that each valuable has a closed-form solution, and discrete hash codes can be obtained directly. The whole optimal algorithm of EDSH is summarized in Algorithm 1.\par
Since the hash functions of each modality have been learned in the training procedure, the hash code of the query can be generated by the corresponding hash functions and orthogonal rotation matrix, directly. And then the Hamming distances between the query and the candidate data points from different modality can be easily calculated by XOR operation. Finally, those data points with relatively smaller Hamming distance to query data are returned.\par

\begin{algorithm}[t]
\caption{Efficient Discrete Supervised Matrix Factorization Hashing} \label{alg1}
\begin{algorithmic}[1]
\REQUIRE The feature matrix of training data points $\{X^{(1)},X^{(2)}\}$ and their corresponding class label matrix $Y$, and the length of hash codes $k$.
\STATE Initializing the hash codes $B$, the sharing space $V$, the orthogonal rotation matrix $R$, hash functions $W_1$ and $W_2$, randomly. And then using Eq.(10), Eq.(11), Eq.(12) to initialize $U^{(1)}$, $U^{(2)}$, $P$, respectively. \FOR{$i=1$ to $miter$} \STATE Update $U^{(1)}$ with fixing the other valuables by Eq.(10), \STATE Update $U^{(2)}$ by fixing other valuables using Eq.(11), \STATE Update $P$ by fixing other valuables using Eq.(12), \STATE Update the sharing space $V$ by fixing other valuables using Eq.(14), \STATE Update the orthogonal rotation matrix $R$ by fixing other valuables using $R={\hat S}S^T$, \STATE Update the hash codes $B$ by fixing other valuables using Eq.(18), \STATE Update the hash functions of image modality $W_1$ by fixing other valuables using Eq.(21),\STATE Update the hash functions of text modality $W_2$ by fixing other valuables using Eq.(22), \ENDFOR  \ENSURE The hash functions of image and text modality $W_1$, $W_2$ and the orthogonal rotation matrix $R$.
\end{algorithmic}
\end{algorithm}

\subsection{Computational Complexity Analysis}
In this section, we demonstrate that the computational complexity of our EDSH is linear to the size of training set. The training computational complexity of EDSH includes the following parts: the computational complexity of solving Eq.(10), Eq.(11), Eq.(12) are $O(d_1kN+k^2N+k^3)$, $O(d_2kN +k^2N+k^3)$ and $O(ckN+k^2N+k^3)$, respectively; for solving Eq.(14) is $O(d_1 k^2+d_2 k^2+ck^2+2k^3+2d_1 kN+2d_2 kN+ckN+2k^2N)$; for solving Eq.(14) is $O(k^3+k^2N)$; for solving $R$ is $O(k^3+k^2N)$; for solving Eq.(17) is $O(kN)$; for solving Eq.(20) and Eq.(21) are $O(kd_1 N+d_1^2 N+d_1^3+kd_1^2)$ and $O(kd_2 N+d_2^2 N+d_2^3+kd_2^2)$, respectively. The size of training number $N$ is much greater than $k$, $d_1$, $d_2$, $c$ in large-scale applications, thus the computational complexity of each iteration is linear to the training size $N$. Given the number of iterations $T$ , which is typically smaller than 20 in our experiments, the overall training computational complexity of EDSH is $O(N)$.
\section{Experiments}
In this section, we conduct comparison experiments to evaluate the effectiveness and efficiency of the proposed EDSH with several existing cross-modal hashing methods on three real world datasets, \emph{i.e.}, Wiki, Mirflickr25K and NUS-WIDE. All experiments are conducted on a server with Intel(R) Xeon(R) CPU E5-2650 v2@2.6GHz and 128 GB memory.
\subsection{Datasets}
In our experiments, three public datasets are utilized to evaluate the effectiveness and efficiency of the proposed EDSH. \par
\textbf{Wiki dataset:}\cite{Rasiwasia-et-al:scheme} This dataset consists of 2,866 image-text pairs collected from Wikipedia. Each pair is assigned to one of 10 ground-truth semantic concepts. For each image, we encode it with a 4,096-dimensional CNN feature extracted by the Caffe implementation of AlexNet \cite{Krizhevsky2012ImageNet}. For each text, we encode it with a 10-dimensional topics vector. Here we randomly sample 25\% data points as the query set, and the remaining 75\% data points for training.\par
\textbf{Mirflickr25K:}\cite{Huiskes2008The} This dataset contains a total of 25,000 images as well as their corresponding tags collected from Flickr. Each pair belongs to more than one of 24 ground-truth semantic concepts. Following \cite{Lin2015Semantics}, only the tags appearing at least 20 times are kept and drop those pairs which have not textual tags or class labels. Accordingly, there are 20,015 pairs in our experiments. Similar to the Wiki dataset, each image is encoded with the 4,096-dimensional CNN vector extracted by the Caffe implementation of AlexNet. We randomly select 80\% data points to generate the query set, and the rest as the training set. \par
\textbf{NUS-WIDE:}\cite{Chua-et-al:scheme} The NUS-WIDE dataset contains a total of 269,648 images with a total number of 5,018 tags from 81 semantic concepts. We only choose the top 21 most frequent labels and then 196,776 data points are kept. Each image is represented by a 4,096-dimensional CNN feature vector extracted by the Caffe implementation of AlexNet, and each text is represented by a BOW vector. We take 99\% of the data points and the remaining 1\% data points as the training set and the query set, respectively.
\subsection{Baseline Methods and Implementation Details}
To investigate the retrieval performance of the proposed EDSH, we conduct comparison experiments with several existing cross-modal hashing methods including PDH \cite{Rastegari2013Predictable}, SCM-S \cite{Zhang-Li:kl-one}, CMFH\cite{Ding-et-al:scheme}, LSSH \cite{Zhou-et-al:scheme}, DASH\cite{dekui-et-al:scheme}, SMFH\cite{Tang2016Supervised}, FSH\cite{Liu2017Cross} and DLFH\cite{TIP2019JIANG}. For DASH, the authors propose to firstly learn hash codes by ITQ \cite{Gong-Lazebnik:kl-one} for one modality, and then linear regression is applied to map the other modality to the learned hash codes. Depending which modality is fist used to generate hash codes, we denote them as DASHi (image modality) and DASHt (text modality). Moreover, among them, the PDH, SCM, SMFH, DASH and DLFH explore class labels to improve the retrieval performance, thus they are supervised methods and the rest of them are unsupervised ones. All source codes of baseline methods are kindly provided by the authors. In our experiments, all parameters of baseline methods are chosen according to their original paper. Moreover, due to the high training time on all training data points, we randomly select 5,000 data points to train hash functions for SMFH on NUS-WIDE dataset as \cite{dekui-et-al:scheme} and \cite{Ding-et-al:scheme} do.\par
The parameters of EDSH are chosen by a cross validation phase. In our experiments, we set $\lambda_1=1$, $\lambda_2=1$, $\gamma=10$, $\alpha=2$, $\beta_1=10$, $\beta_2=10$ and $\mu=5$. To reduce the randomness of initializations and data points selection, all experiments are run 5 times, and we report the averaged experimental results.\par
\subsection{Evaluation Criteria}
In this paper, three criteria, \emph{i.e.}, Mean Average Precision (MAP), Top-k precision curve and PR (Precision-Recall) curve, which are widely used in multi-media retrieval domain, are adopted to evaluate the retrieval performance from different aspects. Given a group of query data points ranked at top in the rank list, the average precision (AP) for each query data point is defined as following\par
\begin{equation}AP=\frac{1}{L}\sum_{m=1}^M P(m)* \delta(m)\end{equation}
where $M$ is the number of returned data points, $L$ denotes the total number of semantic related data points to the query, $P(m)$ denotes the precision value of the top $m$ returned data points, and $\delta(m)=1$ if the $m$-th retrieved data point is the true neighbor of the query data, and $\delta(m)=0$ otherwise. The MAP is the average of AP values for all query data points. Besides, Top-k precision is defined as the precision on top $k$ returned data points, and Precision-Recall (PR) curve is defined as the precision at different recall ratios.\par

\subsection{Experimental Results}

\begin{table*}[ht]
\setlength{\abovecaptionskip}{6pt}
\setlength{\belowcaptionskip}{8pt}
\label{tab:Tab1_discrision_dataset}
\begin{center}
\caption{The mAP@100 scores comparison on Wiki, Mirflickr25K and NUS-WIDE datasets} 
{\renewcommand\arraystretch{1.4}
\fontsize{9pt}{12pt}\selectfont
\setlength{\tabcolsep}{1.4mm}{
\begin{tabular}{c|c|c|c|c|c|c|c|c|c|c|c|c|c}
\hline
\multirow{2}{*}{Task} &\multirow{2}{*}{Methods} & \multicolumn{4}{|c|}{Wiki}  & \multicolumn{4}{|c|}{Mirflickr25K} & \multicolumn{4}{|c}{NUS-WIDE}\\%
\cline{3-14}
& & $8$ &$16$  &$24$ &$32$ &$8$ &$16$  &$24$ &$32$ & $8$ &$16$  &$24$ &$32$\\
\hline																	
\multirow{10}{*}{ }
&PDH \cite{Rastegari2013Predictable} & 0.2023  &0.2051  &0.2070  &0.2174  &0.6240  &0.6231  &0.6345  &0.6423  &0.4439& 0.4458& 0.4311 &0.4612\\
 Image&SCM-S\cite{Zhang-Li:kl-one}   &0.2417 & 0.2672  &0.2724   &0.2830    & 0.8241 &0.8692  &0.8729&0.8766&0.3325 & 0.3414 &0.3578   &0.3610\\
&CMFH\cite{Ding-et-al:scheme}        & 0.2180  &0.2181  &0.2189  &0.2283  &0.6331  &0.6357  &0.6391  &0.6412  &0.4039  &0.4260  &0.4378  &0.4426\\
to&LSSH\cite{Zhou-et-al:scheme}      &0.2212  &0.2224  &0.2293  &0.2318  &0.6228  &0.6324  &0.6430  &0.6515  &0.4720  &0.4877  &0.4895  &0.4989 \\
&SMFH\cite{Tang2016Supervised}       &0.2331  &0.2446  &0.2618 &0.2480&0.6349  &0.6423  &0.6425  &0.6470  &0.3357 &0.3369 &0.3417  &0.3473 \\
Text&DASHi\cite{liang2016online}    &0.2748 &0.2955 &0.3058  &0.3072 & 0.8203 &0.8514 &0.8521& 0.8657 &0.4610 &0.5104 &0.5130 &0.5188 \\
&DASHt\cite{liang2016online}         &0.2614&0.2865 &0.2827& 0.3108 & 0.8025&0.8290&0.8368& 0.8369&0.4737 &0.5240&0.5329&0.5485\\
&FSH\cite{Liu2017Cross}              &0.2347 &0.2440 &0.2613 &0.2618     &0.5324 &0.5539 &0.5580&0.5890 &0.4294 &0.4681 &0.4838 &0.4988 \\
&DLFH\cite{TIP2019JIANG}          & 0.1755  &0.2555 &0.2808  &0.3016   & 0.8203 &0.8840 &0.8871 &0.8904 &0.4760 &0.5439 &0.5540& 0.5632 \\
\cline{2-14}
&EDSH   &\textbf{0.4517} &\textbf{0.4777} &\textbf{0.4833} &\textbf{0.4704} &\textbf{0.8958} &\textbf{0.9009} &\textbf{0.8994}&\textbf{0.9013}  &\textbf{0.5168} &\textbf{0.5892} &\textbf{0.5925} &\textbf{0.5977}\\

\hline
\multirow{10}{*}{ }
&PDH \cite{Rastegari2013Predictable}  &0.3550 & 0.3706& 0.3653 &0.3700  &0.7298 & 0.7283& 0.7613 &0.7920 &0.4144  &0.4195  &0.4032  &0.4040 \\
Text&SCM-S\cite{Zhang-Li:kl-one}      &0.6305 &0.6344   &0.6352  &0.6359  &0.8420  &0.8739  &0.8756 &0.8874 &0.3449 &0.3605  &0.3676& 0.3768  \\
&CMFH\cite{Ding-et-al:scheme}        & 0.5253  &0.5182  &0.5350  &0.5447 &0.6315 &0.6387 &0.6479  &0.6406  &0.6811  &0.7295  &0.7366  &0.7387\\
to&LSSH\cite{Zhou-et-al:scheme}      &0.5981  &0.6115 &0.6207  &0.6169  &0.7093 &0.7388  &0.7460   &0.7593  &0.5910  & 0.6135   & 0.6383   &0.6688 \\
&SMFH\cite{Tang2016Supervised}       &0.6366 &0.6520  &0.6596 &0.6603&0.5694  &0.5762 &0.5858 &0.5835  &0.3651  &0.3686  &0.3754  &0.3885 \\
Image&DASHi\cite{liang2016online}     &0.6214   &0.6508&0.6609  &0.6627 &0.8280 & 0.8632&0.8666& 0.8707 &0.4272 &0.4764 &0.4840  &0.4875\\
&DASHt\cite{liang2016online}         &0.5973   &0.6166 &0.6230 &0.6241 & 0.8273 &0.8605&0.8640&0.8732 &0.4054  &0.4520 &0.4539 &0.4672\\
&FSH\cite{Liu2017Cross}              &0.5520 &0.5734 &0.5909 &0.5875    &0.5352 &0.5576&0.5608&0.5690 &0.4295 &0.4368 &0.4472 & 0.4483\\
&DLFH\cite{TIP2019JIANG}         & 0.4595  &0.6520 &0.6633  &0.6675  &0.7912  &0.8589  &0.8590& 0.8646 &0.7781  &0.8425 &0.8770 &0.8908\\
\cline{2-14}
&EDSH  &\textbf{0.6751} &\textbf{0.6810}&\textbf{0.6719}&\textbf{0.6752} &\textbf{0.9290} &\textbf{0.9406} &\textbf{0.9324}&\textbf{0.9376} &\textbf{0.8536}  &\textbf{0.9059}& \textbf{0.9163}  &\textbf{0.9217}\\
\hline

\end{tabular}}
}
\end{center}
\end{table*}

\subsubsection{Retrieval Performance}
The MAP performance of our EDSH and baseline methods on Wiki, Mirflickr25K and NUS-WIDE for the two cross-modal retrieval tasks are summarized in Table 1. From Table 1, we have the following observations:\par
(1) Our proposed EDSH outperforms all baseline methods in all cases on the three datasets with different code lengths, which shows its effectiveness. Specifically, EDSH achieves superior performance to the best of the baseline methods with performance gains of 35\%, 33\%, 30\%, 27\% on Wiki dataset with the code lengths varying from 8 bits to 32 bits; For the Mirflickr25K dataset, EDSH achieves superior performance to the best of the baseline methods with performance gains of 10\%, 8\%, 7\%, 6\% with the code lengths varying from 8 bits to 32 bits; For the NUS-WIDE dataset, EDSH achieves superior performance to the best of the baseline methods with performance gains of 10\%, 8\%, 7\%, 6\% with the code lengths varying from 8 bits to 32 bits. The superiority of EDSH can be mainly attributed to its capability of better preserving the similarities in hash codes and the discrete optimization algorithm.\par
(2) Generally, the supervised hashing methods can obtain better MAP performance, \emph{i.e.}, SCM-S, DASHi, DASHt and DLFH. The main reason is that supervised methods can explore the class labels to enhance the semantic information in hash codes which makes them more discriminative.\par
(3) The MAP performance of the supervised method SMFH is low on NUS-WIDE dataset. The reason is that we randomly select 5,000 data points to train hash functions due to the high training time cost on large-scale dataset, while other methods learn hash functions on all available training data points.\par

Fig.2 illustrates the Precision-Recall curves in the case of 16 bits code length on the three real world datasets for the two tasks. From this figure, we can observe that EDSH generally achieves much better experimental results than that of the baseline methods in most cases, which well demonstrates the superiority of our proposed method. More specifically, EDSH achieves best performance compared to all baseline methods consistently when the return ratio is relatively small. This is essential for a search engine since users typically pay more attention to those data points ranked higher.\par

Moreover, the Top-K curves on the three real world datasets for the two tasks with the code length fixed to be 16 bits are shown in Fig.3. From this figure, it can be seen that the our EDSH consistently performs best in all cases and performs much better in some cases. This phenomenon is consistent with the MAP performance reported in Table 1.

\begin{figure*}[htbp]
\centering
\subfigure[]{
\label{fig4_11a} 
\includegraphics[width=0.31\textwidth{},keepaspectratio]{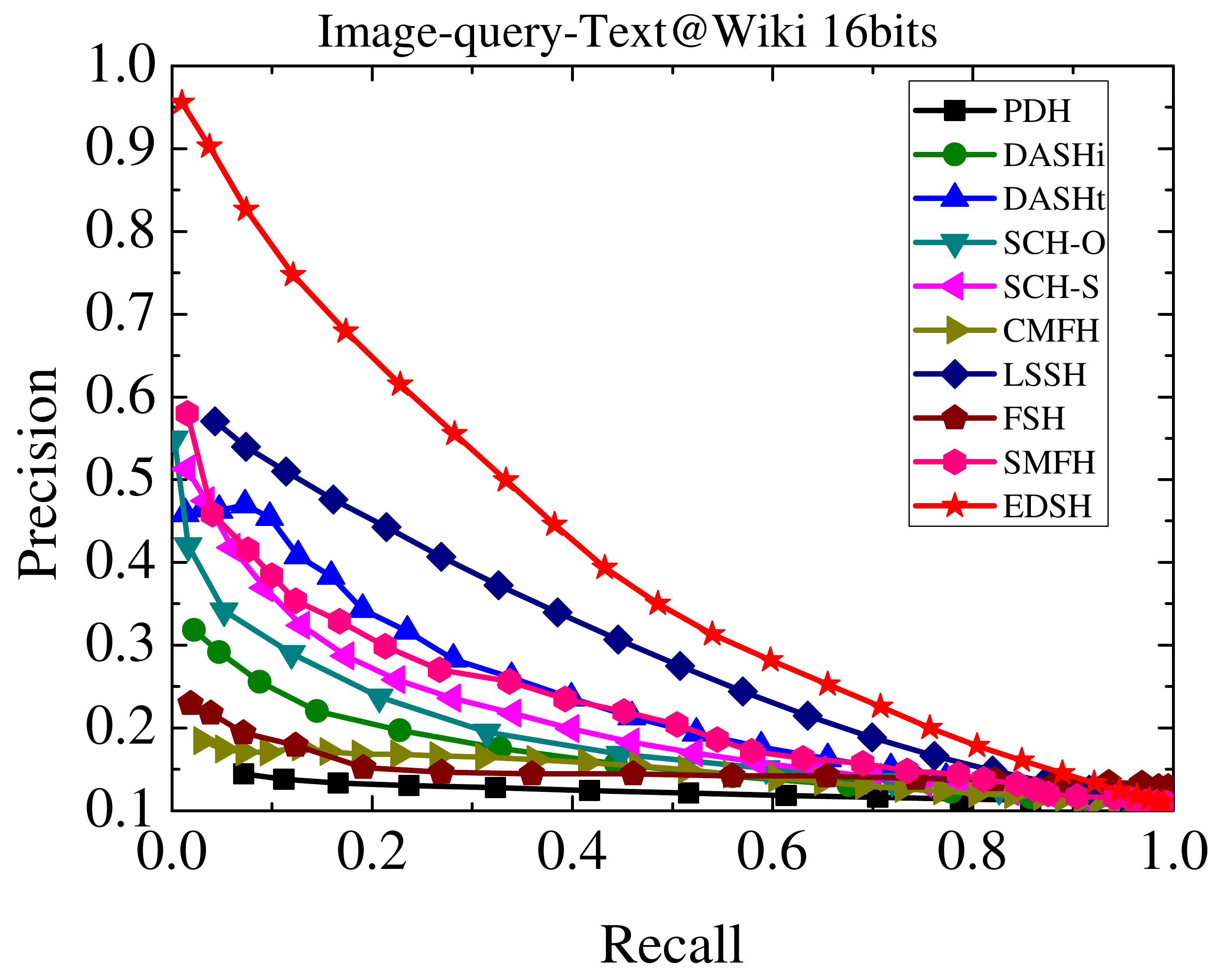}}
\hspace{0in}  
\subfigure[]{
\label{fig4_11b} 
\includegraphics[width=0.31\textwidth{},keepaspectratio]{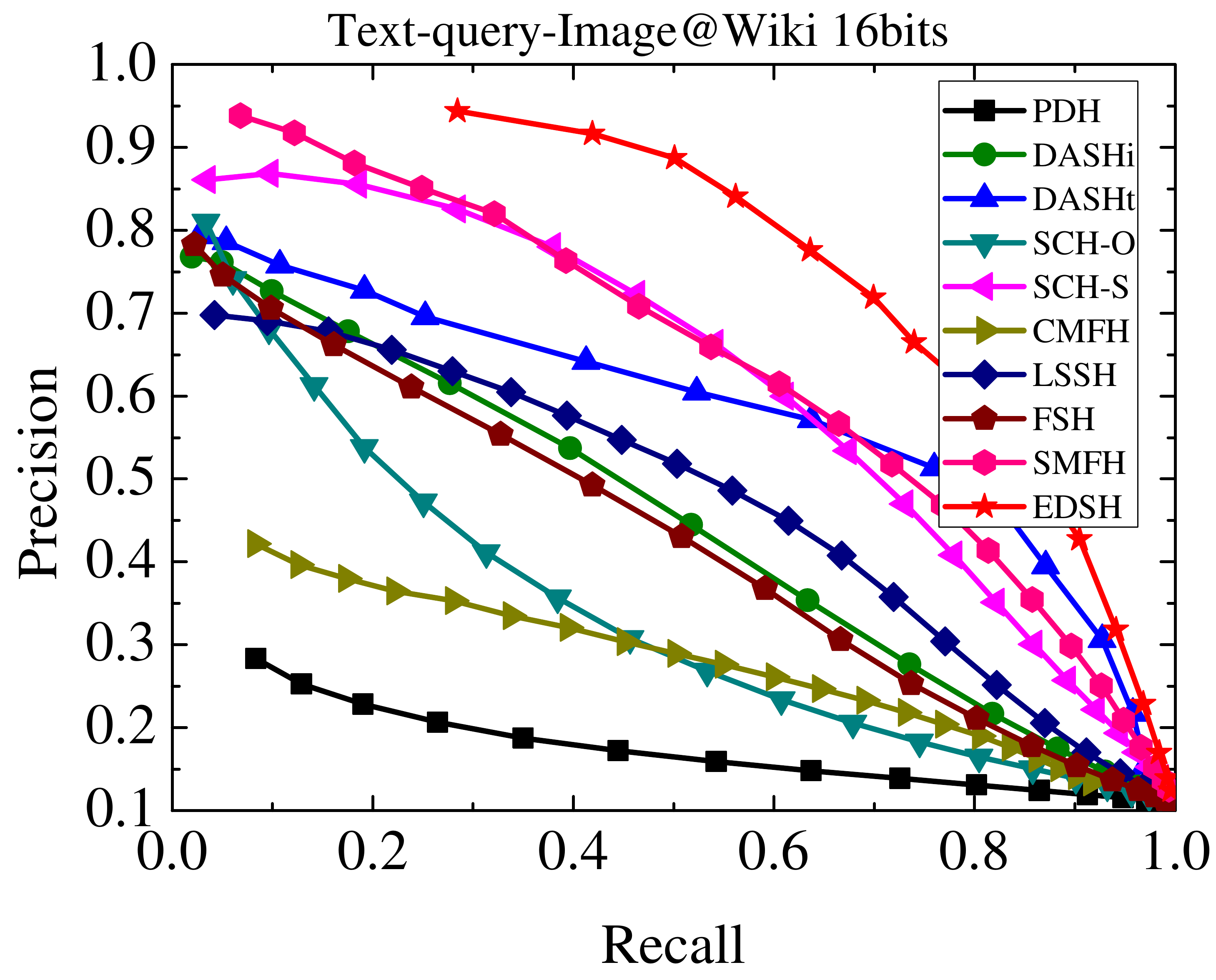}}
\hspace{0in}
\subfigure[]{
\label{fig4_11a} 
\includegraphics[width=0.31\textwidth{},keepaspectratio]{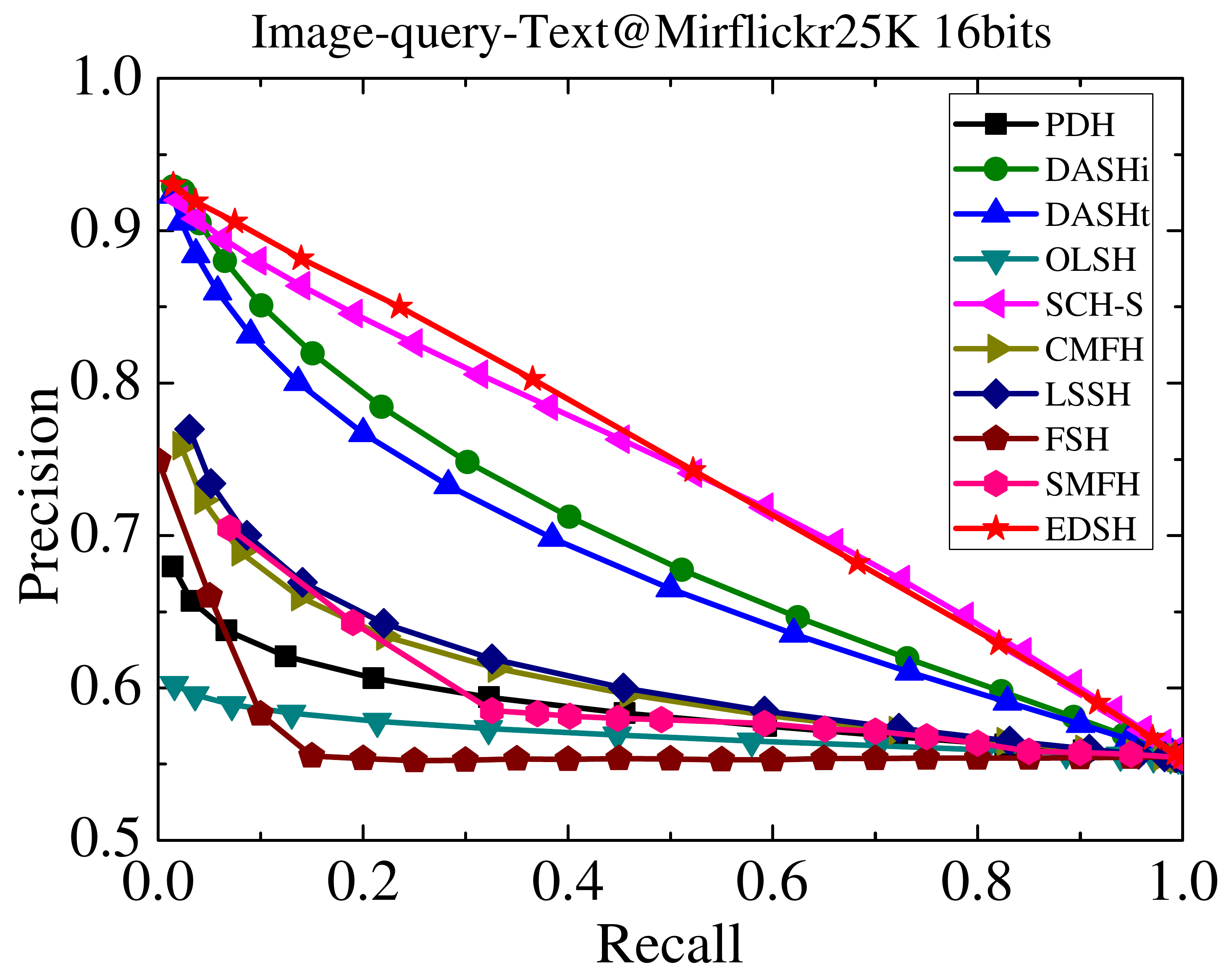}}
\hspace{0in}  
\subfigure[]{
\label{fig4_11b} 
\includegraphics[width=0.31\textwidth{},keepaspectratio]{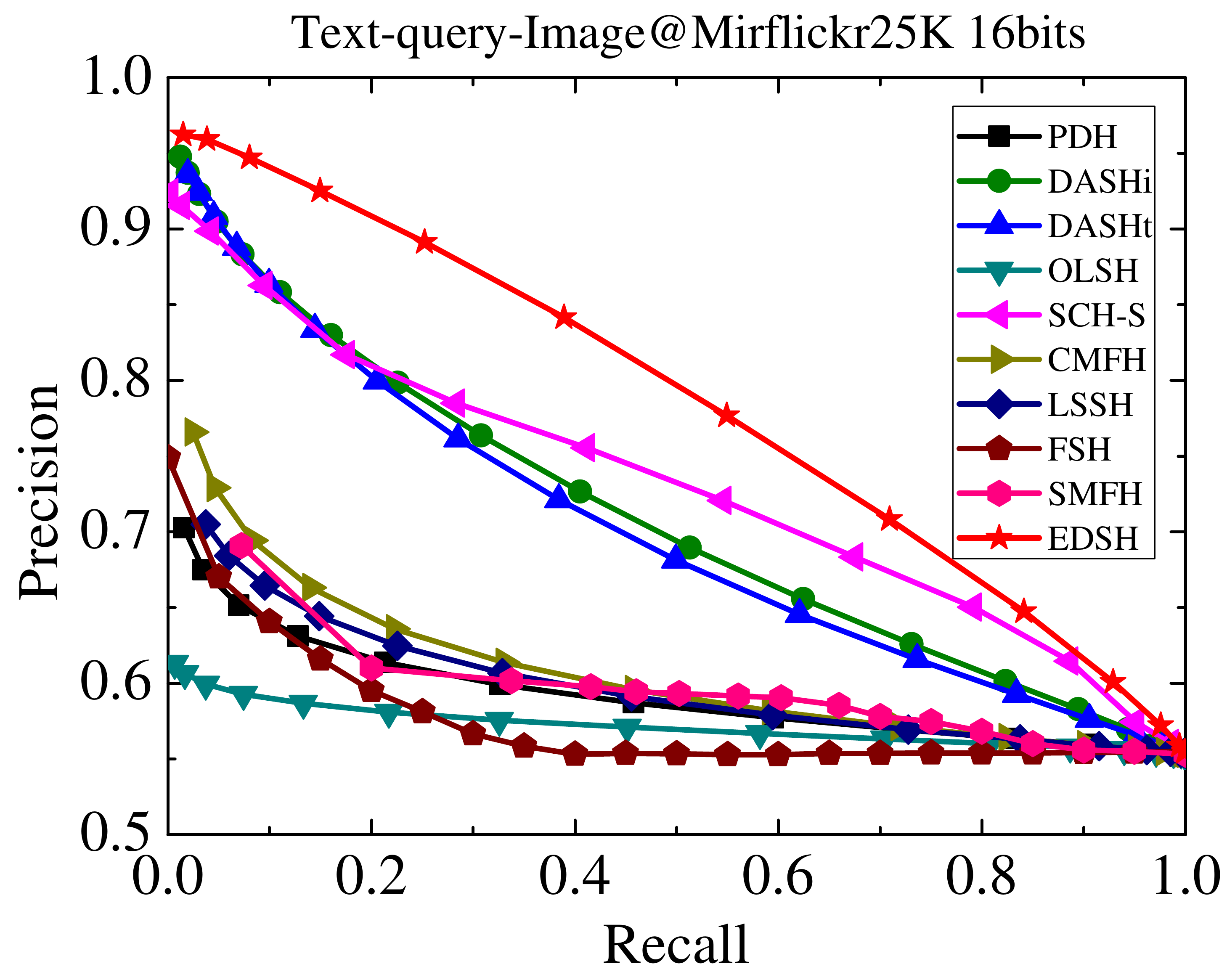}}
\hspace{0in}
\subfigure[]{
\label{fig4_11b} 
\includegraphics[width=0.31\textwidth{},keepaspectratio]{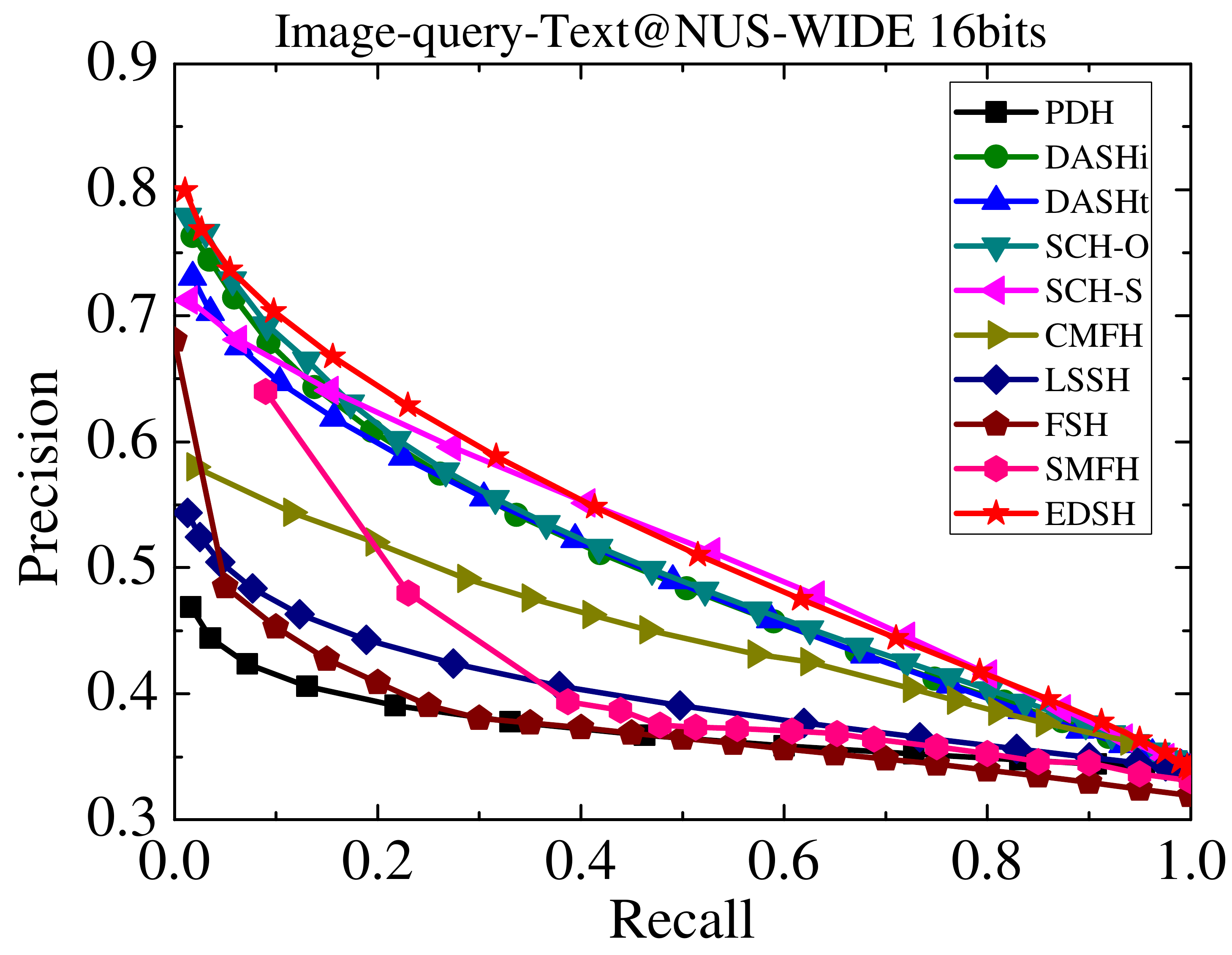}}
\hspace{0in}
\subfigure[]{
\label{fig4_11b} 
\includegraphics[width=0.31\textwidth{},keepaspectratio]{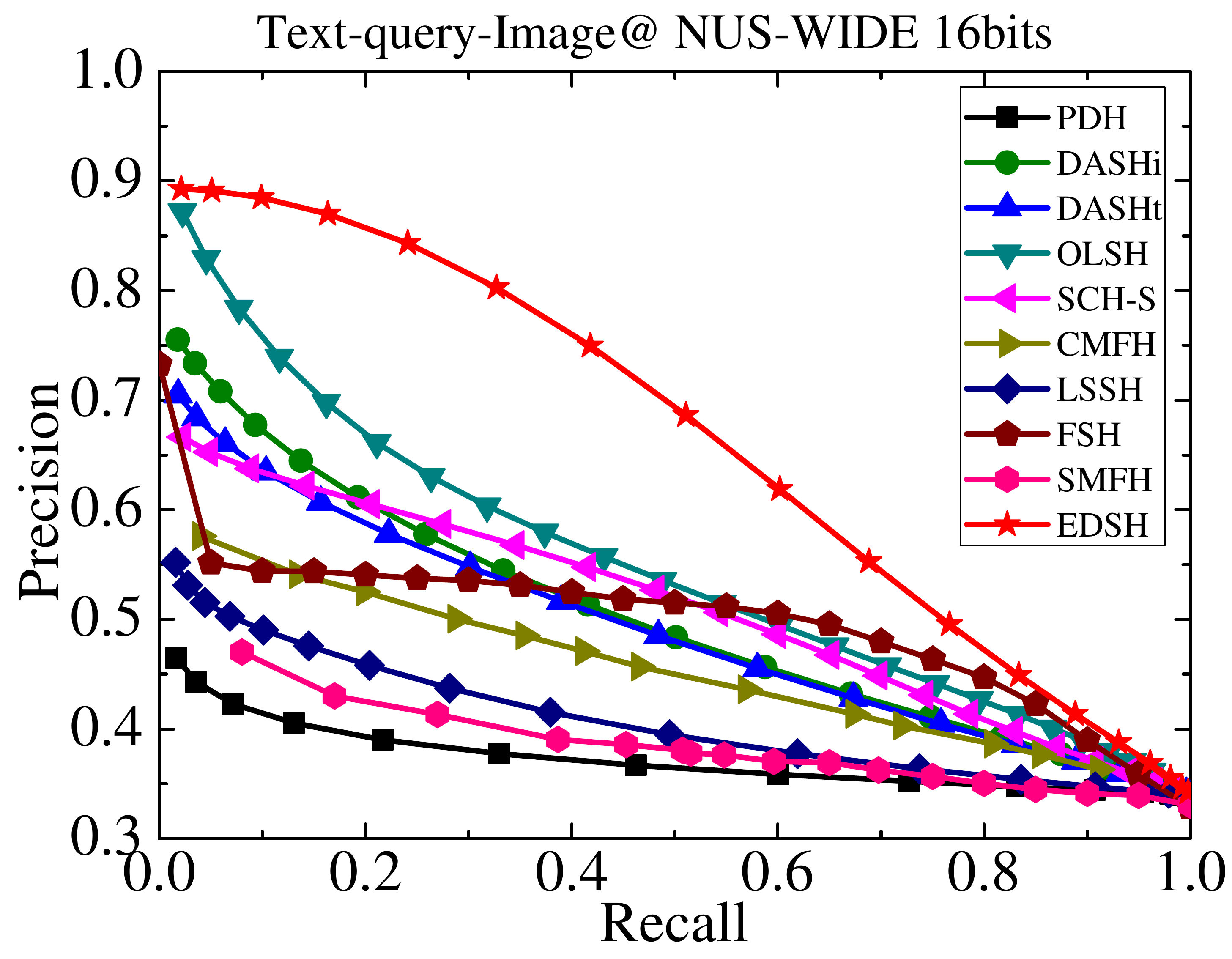}}
\hspace{0in}
\begin {center}
\caption{Precision-Recall curves of EDSH and the baseline methods on Wiki, Mirflickr25K and NUS-WIDE datasets when hash code is 16 bits.}
\end {center}
\end{figure*}

\begin{figure*}[htbp]
\centering
\subfigure[]{
\label{fig4_11a} 
\includegraphics[width=0.31\textwidth{},keepaspectratio]{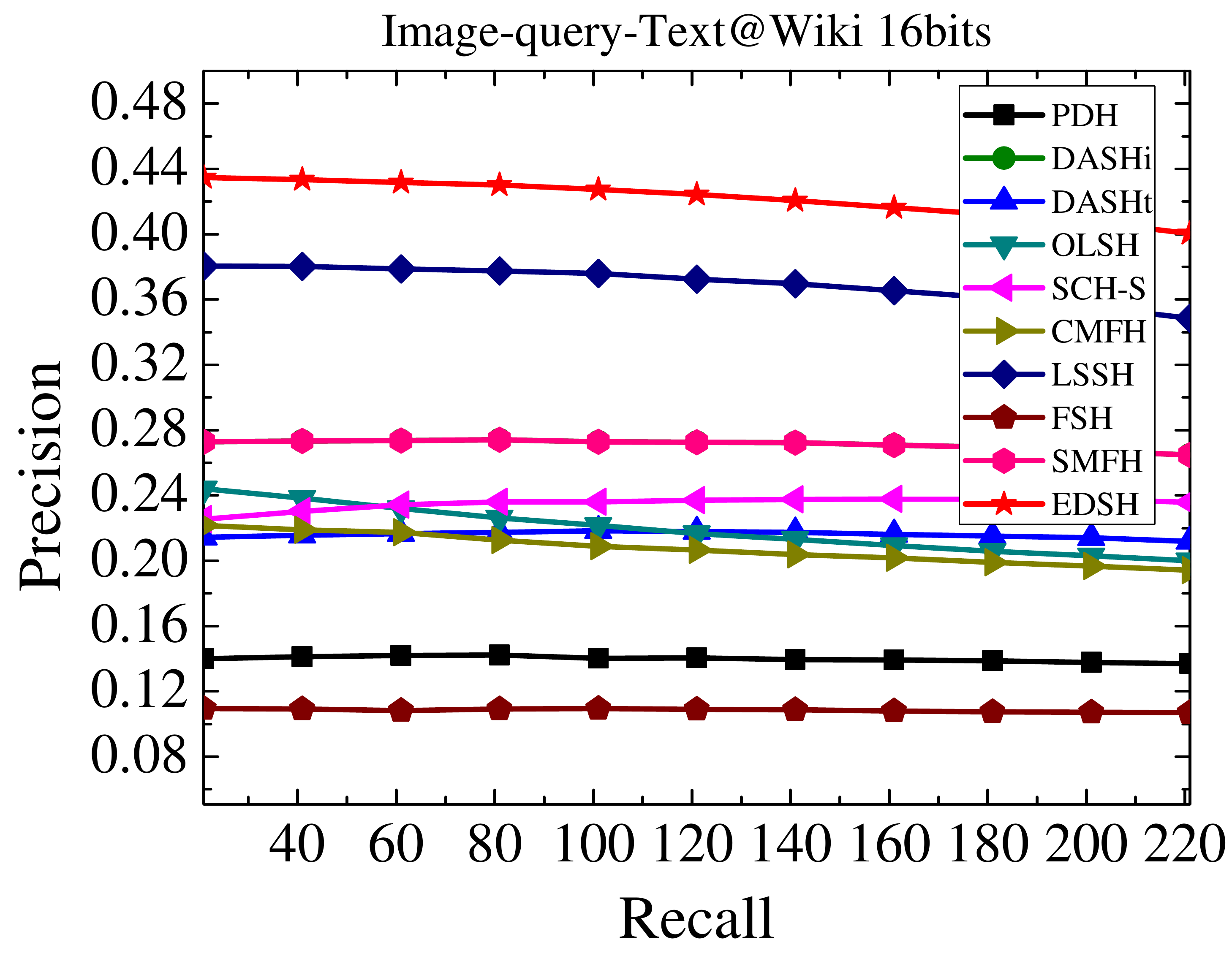}}
\hspace{0in}  
\subfigure[]{
\label{fig4_11b} 
\includegraphics[width=0.31\textwidth{},keepaspectratio]{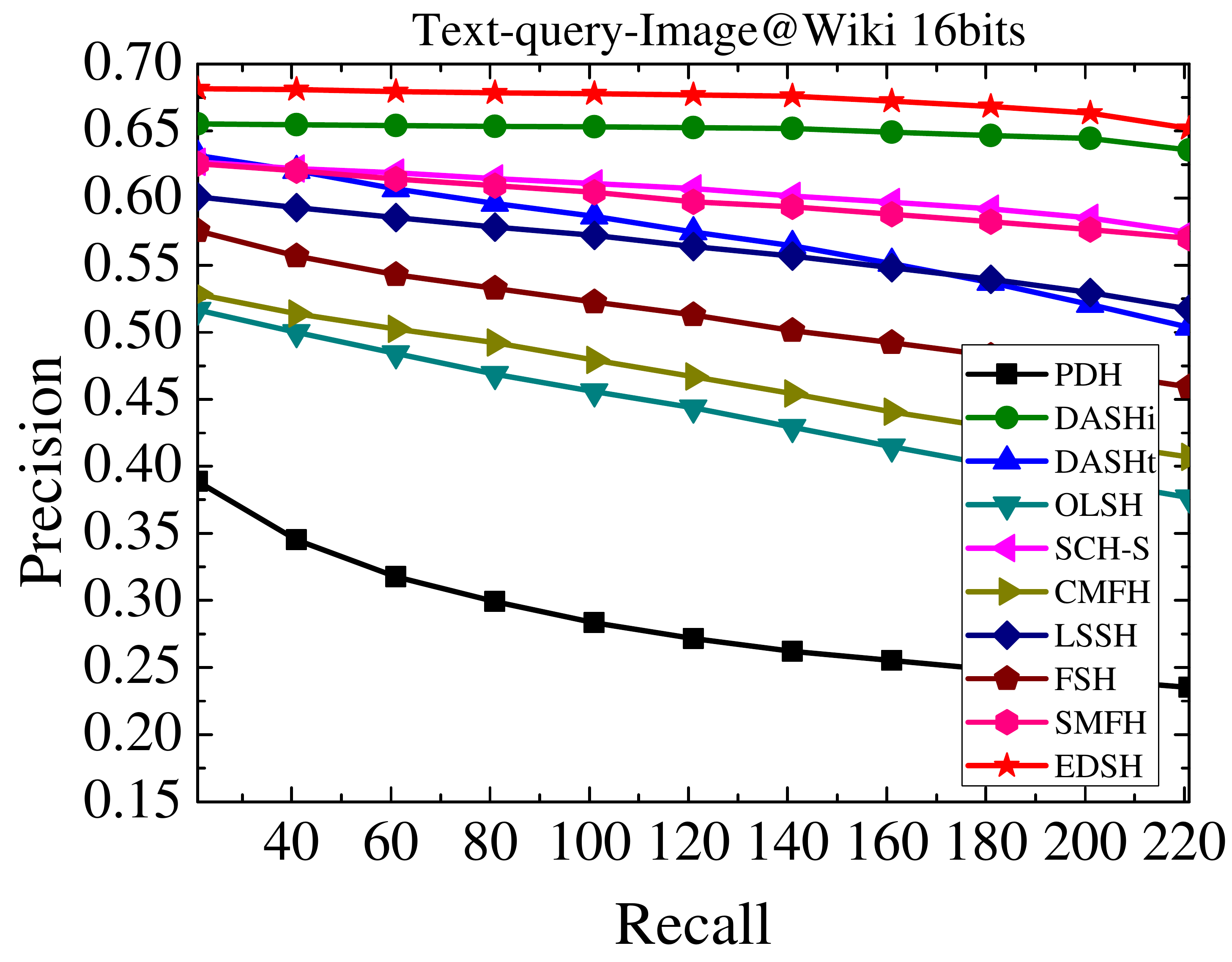}}
\hspace{0in}
\subfigure[]{
\label{fig4_11a} 
\includegraphics[width=0.31\textwidth{},keepaspectratio]{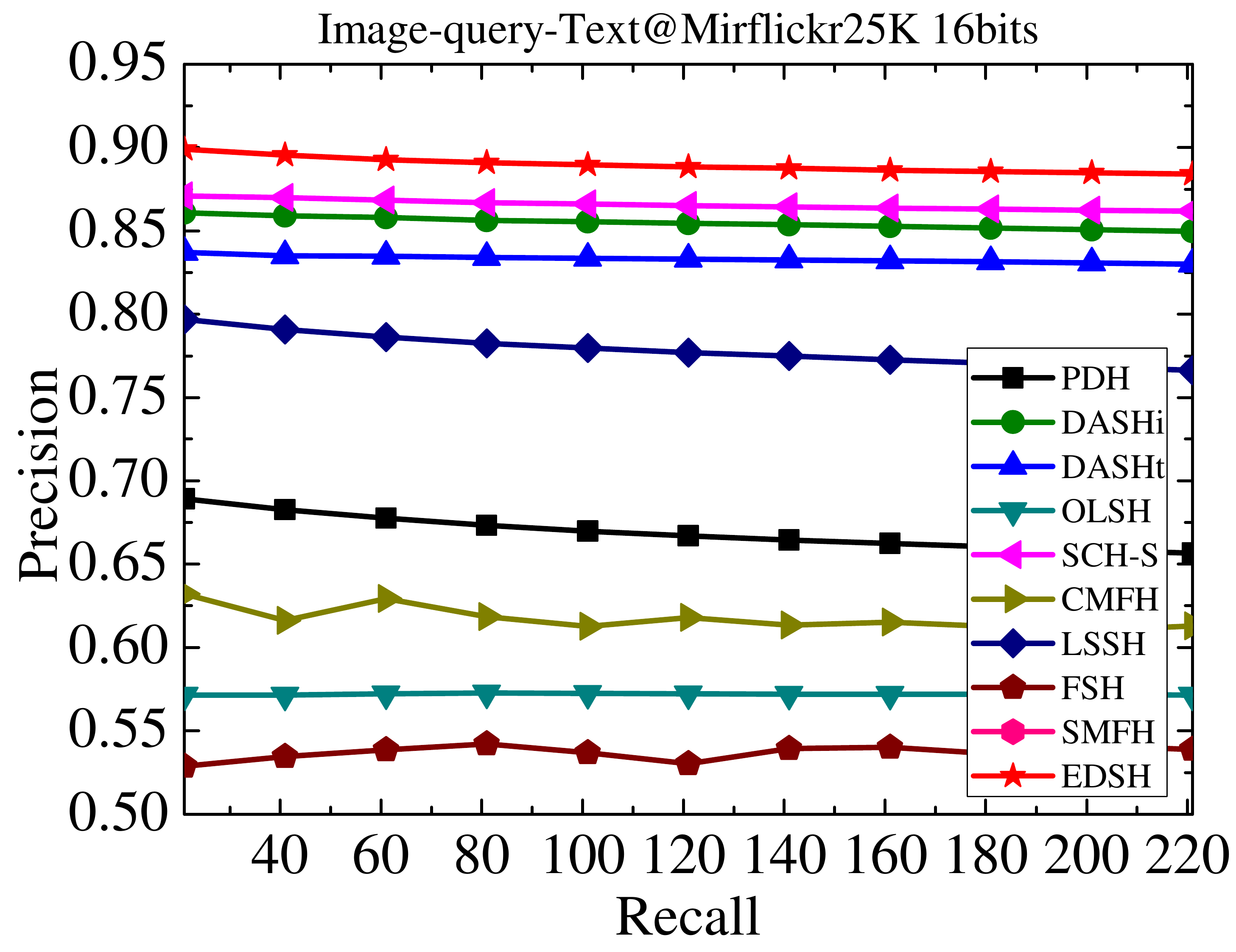}}
\hspace{0in}  
\subfigure[]{
\label{fig4_11b} 
\includegraphics[width=0.31\textwidth{},keepaspectratio]{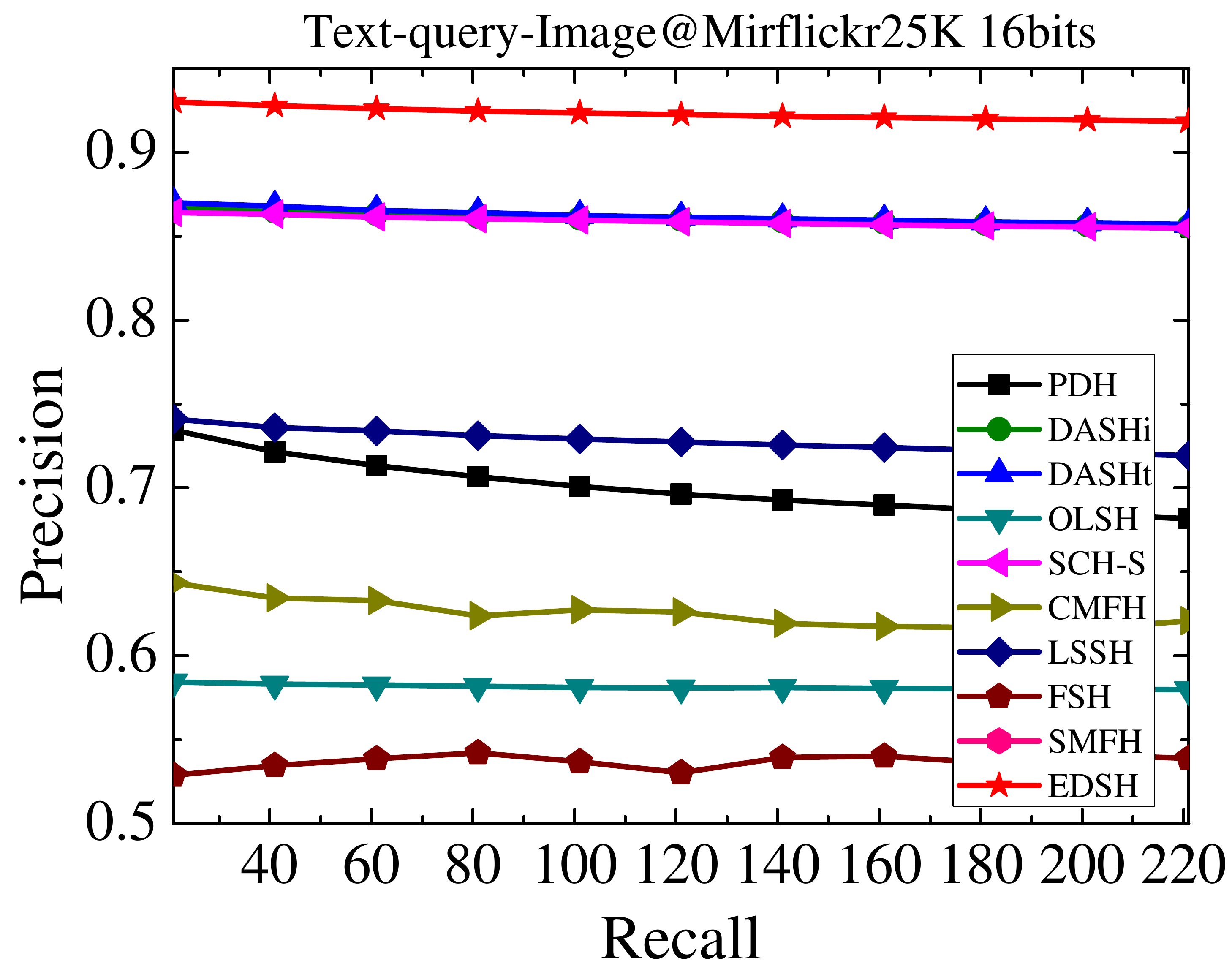}}
\hspace{0in}
\subfigure[]{
\label{fig4_11b} 
\includegraphics[width=0.31\textwidth{},keepaspectratio]{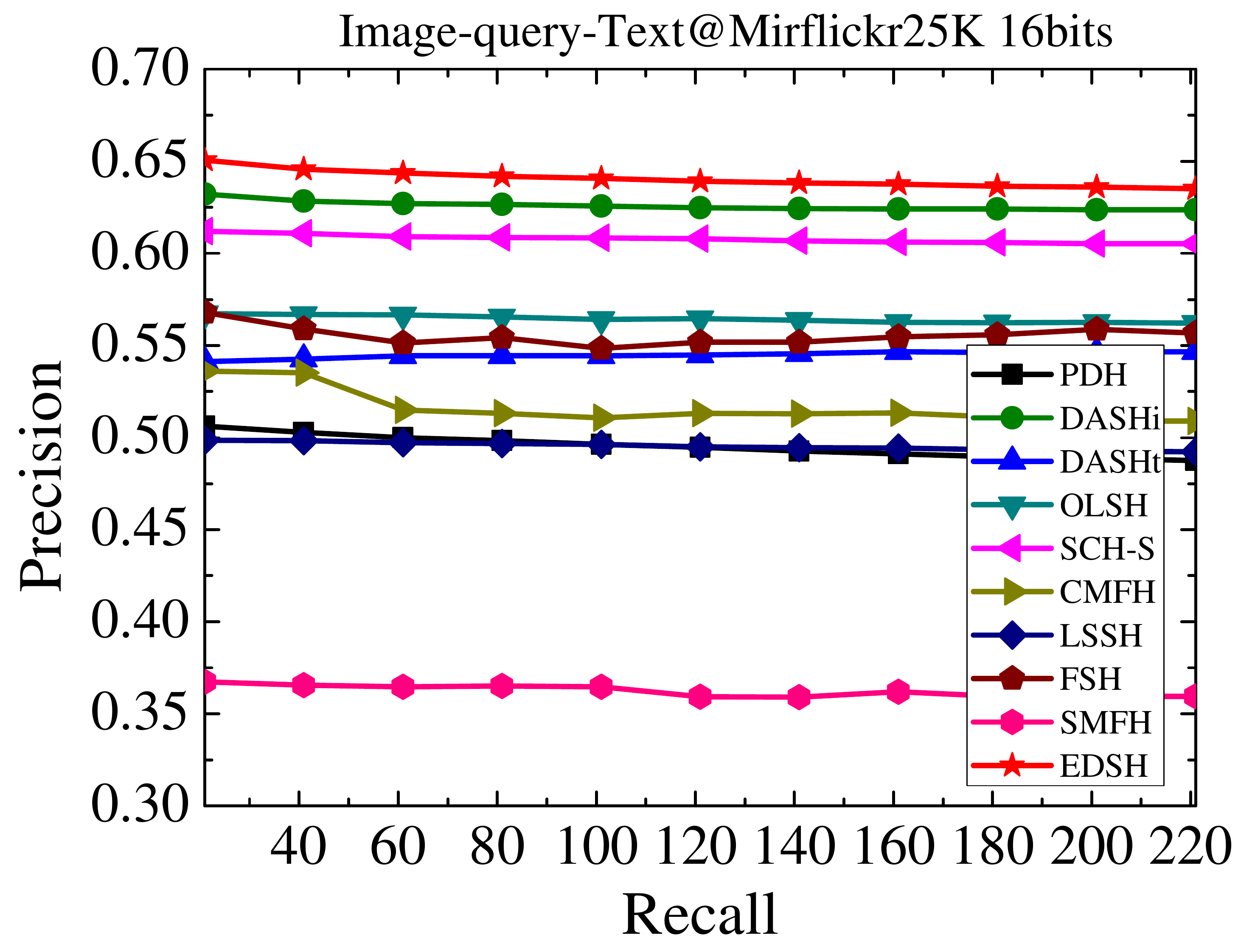}}
\hspace{0in}
\subfigure[]{
\label{fig4_11b} 
\includegraphics[width=0.31\textwidth{},keepaspectratio]{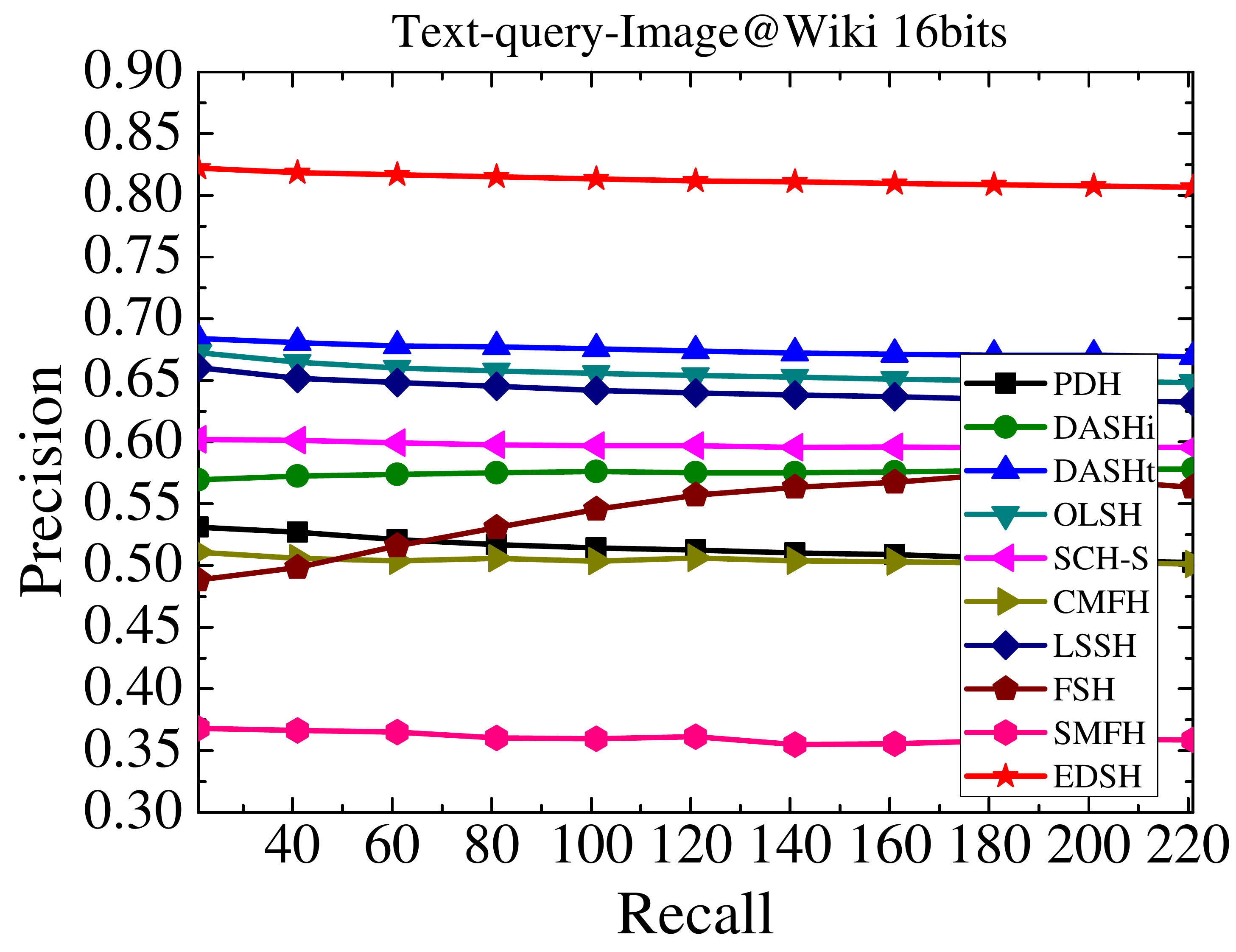}}
\hspace{0in}
\begin {center}
\caption{Top-K curves of EDSH and the baseline methods on Wiki, Mirflickr25K and NUS-WIDE datasets when hash code is 16 bits.}
\end {center}
\end{figure*}

\begin{figure*}[htbp]
\centering
\subfigure[]{
\label{fig4_11a} 
\includegraphics[width=0.3\textwidth{},keepaspectratio]{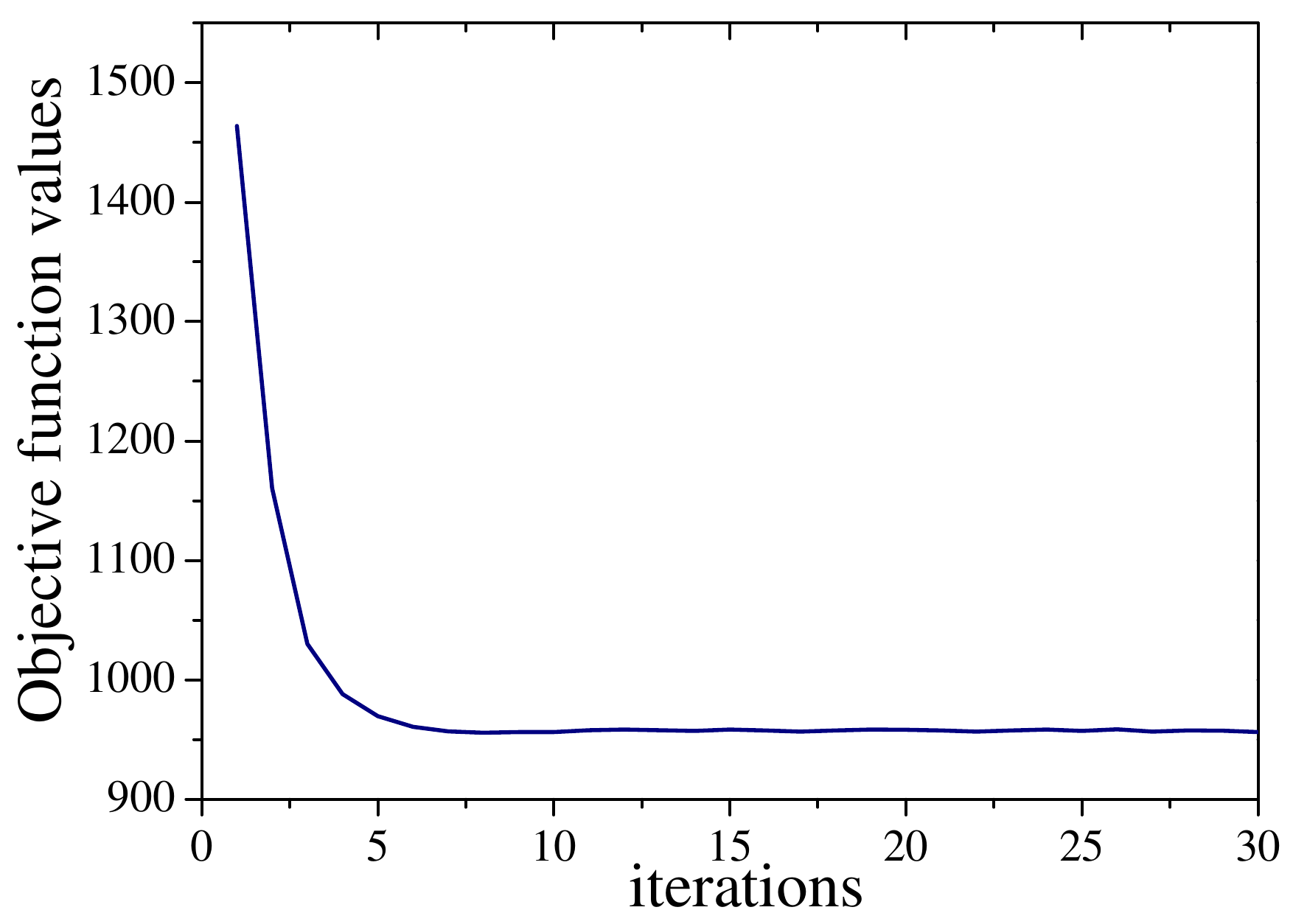}}
\hspace{0in}  
\subfigure[]{
\label{fig4_11b} 
\includegraphics[width=0.3\textwidth{},keepaspectratio]{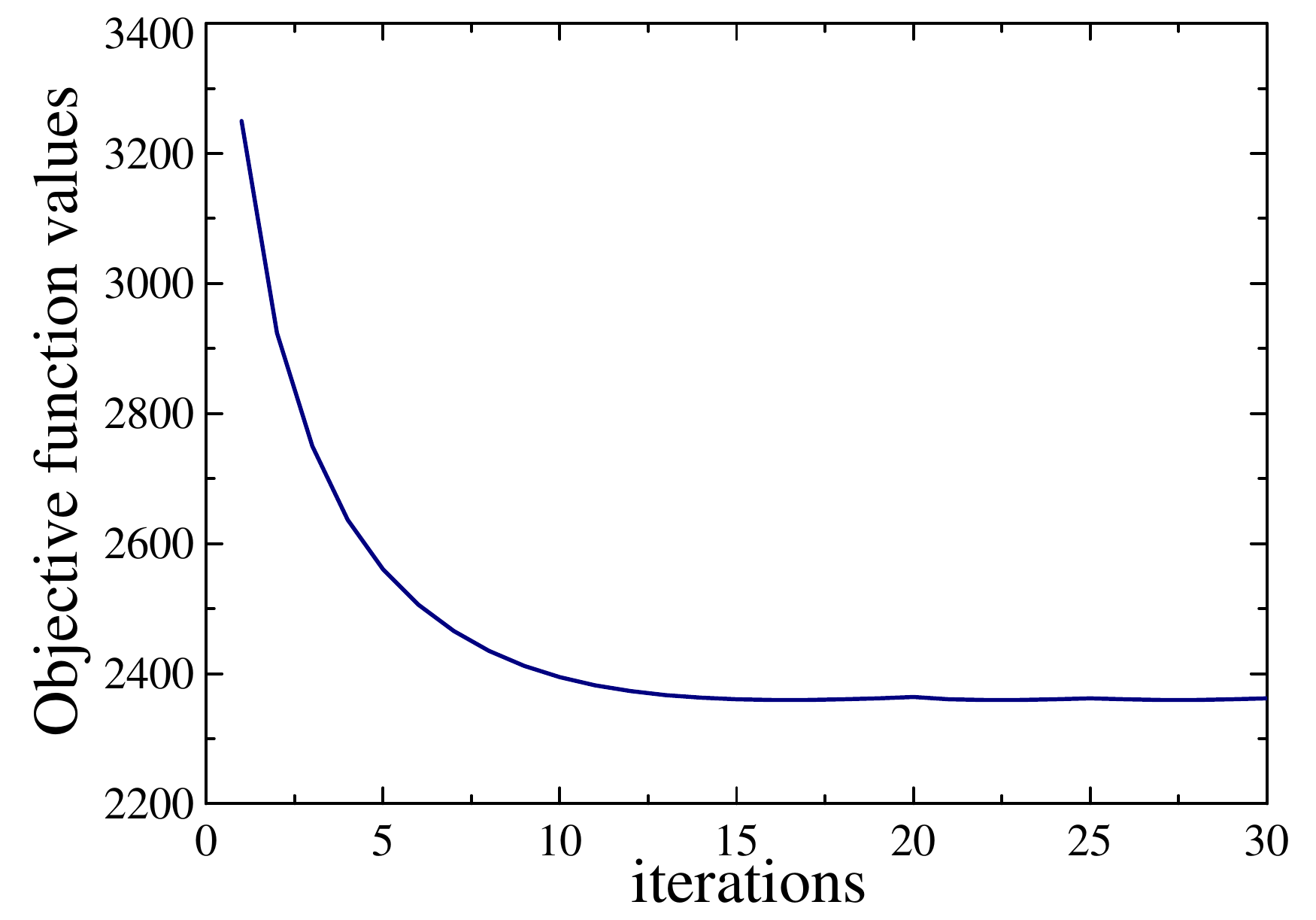}}
\hspace{0in}
\subfigure[]{
\label{fig4_11b} 
\includegraphics[width=0.3\textwidth{},keepaspectratio]{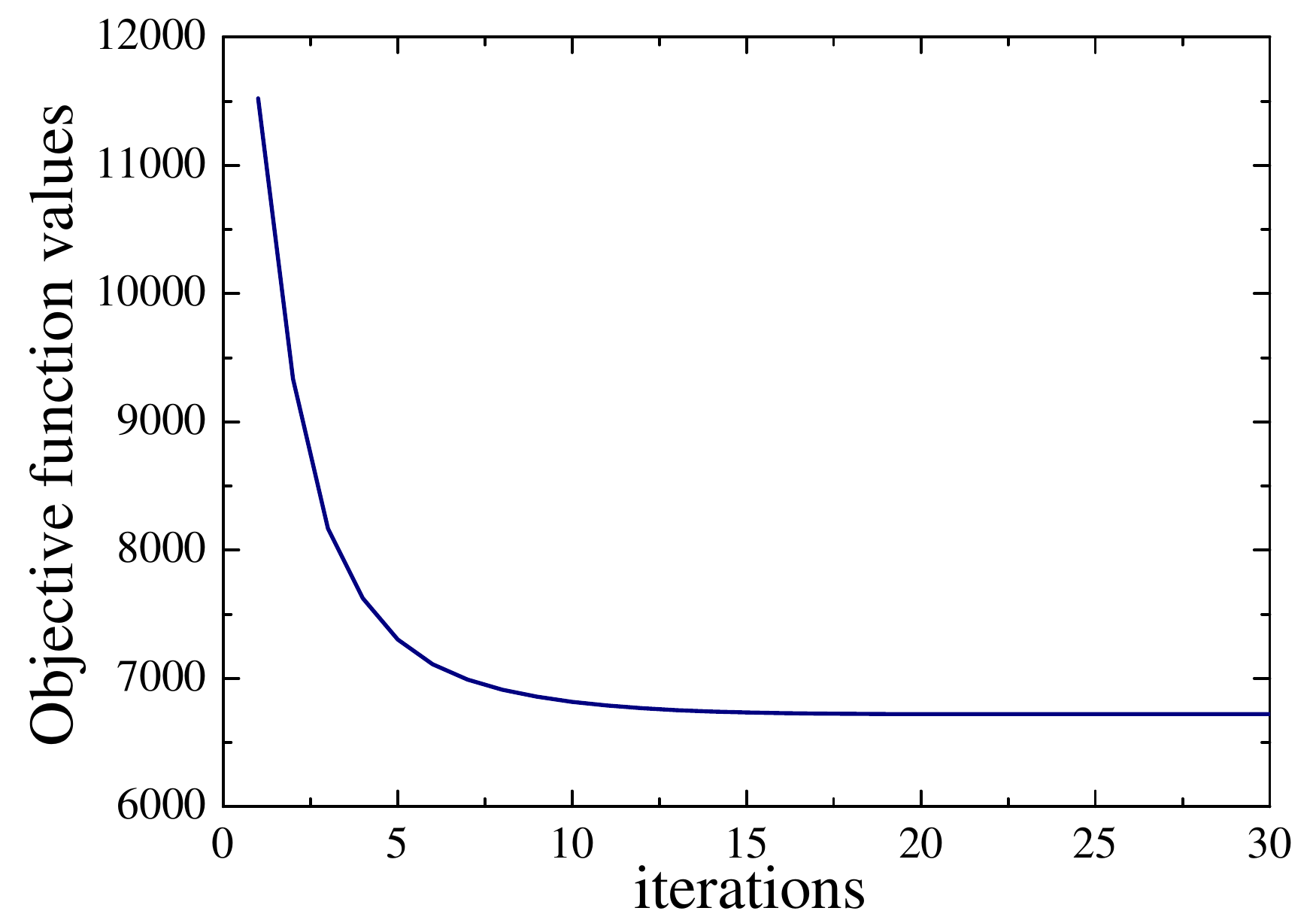}}
\hspace{0in}
\begin {center}
\caption{Convergency Analysis on Wiki, Mirflickr25K and NUS-WIDE datasets.}
\end {center}
\end{figure*}
\subsubsection{Convergency Analysis}
Since the optimal solutions of EDSH are obtained by iterative updating rules, the training time is closely related to the number of iterations in training phase. To demonstrate the efficiency of the proposed method, we show the convergence curves on the three datasets in Fig.4. It can be observed that our EDSH has quick convergence speed on the three datasets in experiments, typically less than 20 iterations, which shows the effectiveness and efficiency of the proposed discrete optimal algorithm.
\subsubsection{Training Time}
The training time of our EDSH and baseline methods on the three datasets are reported in Table 2. Since SMFH costs much time in training procedure on NUS-WIDE dataset, we do not report the training time of it in our experiments. From Table 2, it can be seen that our EDSH costs relatively less time in training phase than most baseline methods. Moreover, our EDSH always performs best on the three datasets compared with other matrix factorization based methods \emph{i.e.}, LSSH, CMFH and SMFH. This demonstrates the efficiency of our method. Although DLFH and DASHt cost less training time than our EDSH on Wiki and Mirflickr25K datasets, but the proposed EDSH costs less time on NUS-WIDE dataset which demonstrates that our EDSH can deal with large-scale datasets more efficient. Therefore, our EDSH possesses not only better retrieval performance but also a competitive computational speed compared with baseline methods.\par
\begin{table}[htbp]
\renewcommand{\arraystretch}{1}
\setlength{\abovecaptionskip}{0pt}
\setlength{\belowcaptionskip}{10pt}
\label{tab:Tab1_discrision_dataset}
\begin{center}
\caption{Training time (in seconds) comparison on the three datasets for 16 bits}
{\renewcommand\arraystretch{1.4}
\fontsize{9pt}{6pt}\selectfont
\begin{tabular}{p{2cm}<{\centering}|p{1.6cm}<{\centering}|p{1.6cm}<{\centering}|p{1.6cm}<{\centering}}
\toprule
Methods&Wiki&Mirflickr25K&NUS-WIDE \\
\midrule
PDH \cite{Rastegari2013Predictable}      &  39  & 335& 3200\\
SCM-S\cite{Zhang-Li:kl-one}              &19836  &20374   & 22402\\

CMFH\cite{Ding-et-al:scheme}              & 16   & 54   & 492\\
LSSH\cite{Zhou-et-al:scheme}             & 38  &177  &1759\\
DASHi\cite{liang2016online}              & 17 & 25  & 114\\
DASHt\cite{liang2016online}             & 10 & 17 & 103\\
FSH\cite{Liu2017Cross}                   & 1164 & 2752 & 8071\\
SMFH\cite{Tang2016Supervised}            & 228 & 842 &-\\
DLFH\cite{TIP2019JIANG}              & 3  &12  &127 \\
EDSH                                   & 13  &22  &91\\
\bottomrule
\end{tabular}
}
\end{center}

\end{table}
\section{Conclusion}
In this paper, we propose a supervised cross-modal hashing method, \emph{i.e.}, Efficient Discrete Supervised Matrix Factorization Hashing (EDSH). It leverages both collective matrix factorization and semantic embedding with class labels to improve the discrimination of hash codes. An efficient discrete optimal algorithm is proposed to directly learn discrete hash code matrix with closed-form solution instead of learning them bit-by-bit. Experimental results on three real world datasets show superior retrieval performance and efficiency of the proposed method over baseline methods.
\section*{Acknowledgment}
This work is supported by the National Natural Science Foundation of China (Grant No. 61872170, 61172109, 61877061).
\input{bare_jrnl.bbl}

\end{document}